\documentclass[11pt]{article}

\usepackage[preprint]{acl}

\usepackage{times}
\usepackage{latexsym}

\usepackage[T1]{fontenc}
\usepackage[utf8]{inputenc}

\usepackage{microtype}

\usepackage{inconsolata}

\usepackage{graphicx}
\usepackage{subcaption}  
\usepackage{float}  
\usepackage{epstopdf}

\newcommand{\promptspacing}{0.5em}
\usepackage{enumitem}
\usepackage{longtable}
\usepackage{listingsutf8}
\usepackage{booktabs}
\usepackage[table]{xcolor}
\usepackage{dblfloatfix}
\usepackage{soul}
\usepackage{cuted,lipsum}
\usepackage{pifont}% http://ctan.org/pkg/pifont
\newcommand{\cmark}{\ding{51}}%
\newcommand{\xmark}{\ding{55}}%
\usepackage{multirow}
\usepackage{makecell}
\usepackage[normalem]{ulem}
\useunder{\uline}{\ul}{}
\usepackage{caption}
\usepackage{subcaption}

\definecolor{codegreen}{rgb}{0,0.6,0}
\definecolor{codegray}{rgb}{0.5,0.5,0.5}
\definecolor{codepurple}{rgb}{0.58,0,0.82}
\definecolor{backcolour}{rgb}{0.95,0.95,0.92}
\colorlet{orange}{black}
\colorlet{blue}{black}

\lstdefinestyle{mystyle}{
    backgroundcolor=\color{backcolour},   
    commentstyle=\color{codegreen},
    keywordstyle=\color{magenta},
    numberstyle=\tiny\color{codegray},
    stringstyle=\color{codepurple},
    basicstyle=\ttfamily\scriptsize,
    breakatwhitespace=false,         
    breaklines=true,                 
    captionpos=b,                    
    keepspaces=true,                                   
    numbersep=5pt,                  
    showspaces=false,                
    showstringspaces=false,
    showtabs=false,                  
    tabsize=2,
    inputencoding=utf8/latin1,
    extendedchars=true, 
    literate= {á}{{\'a}}1 
    {ø}{{\o}}1 
    {æ}{{\ae}}1 
    {å}{{\aa}}1 
}
\usepackage[most]{tcolorbox}
\tcbuselibrary{breakable,skins}

\newtcolorbox{casebox}[1][]{
  colback=gray!3, 
  boxrule=0.3pt, 
  arc=1pt,
  left=2mm, 
  right=2mm, 
  top=1mm, 
  bottom=1mm,
  enhanced, 
  breakable,
  width=\textwidth,  
  fontupper=\footnotesize,
  #1
}

\title{LLMs as annotators of credibility assessment in Danish asylum decisions: evaluating classification performance and errors beyond aggregated metrics}
\author{
 \textbf{Galadrielle Humblot-Renaux\textsuperscript{1,2}},
 \textbf{Mohammad N. S. Jahromi\textsuperscript{1,3,2}},
 \textbf{Rohat Bakuri-Jørgensen\textsuperscript{1}},
\\
 \textbf{Marieke Anne Heyl\textsuperscript{3}},
 \textbf{Asta S. Stage Jarlner\textsuperscript{3}},
 \textbf{Maria Vlachou\textsuperscript{4}},
 \textbf{Anna Murphy Høgenhaug\textsuperscript{3}},
 \\
 \textbf{Desmond Elliott\textsuperscript{4,2}},
 \textbf{Thomas Gammeltoft-Hansen\textsuperscript{3}},
 \textbf{Thomas B. Moeslund\textsuperscript{1,2}}
\\
 \textsuperscript{1}Visual Analysis and Perception Lab, Aalborg University\quad\textsuperscript{2}Pioneer Center for AI, Denmark
 \\
 \textsuperscript{3}Center of Excellence for Global Mobility Law, University of Copenhagen\\
 \textsuperscript{4}Department of Computer Science, University of Copenhagen
\\
 \small{
   \textbf{Correspondence:} \href{mailto:gegeh@create.aau.dk}{gegeh@create.aau.dk}
 }
}

\begin{document}
\maketitle
\begin{abstract}

Off-the-shelf large language models (LLMs) are increasingly used to automate text annotation, yet their effectiveness remains underexplored for underrepresented languages and specialized domains where the class definition requires subtle expert understanding. We investigate LLM-based annotation for a novel legal NLP task: identifying the presence and sentiment of credibility assessments in asylum decision texts. We introduce RAB-Cred, a Danish text classification dataset featuring high-quality, expert annotations and valuable metadata such as annotator confidence and asylum case outcome. We benchmark 21 open-weight models and 30 system-user prompt combinations for this task, and systematically evaluate the effect of model and prompt choice for zero-shot and few-shot classification. We zoom in on the errors made by top-performing models and prompts, investigating error consistency across LLMs, inter-class confusion, correlation with human confidence and sample-wise difficulty and severity of LLM mistakes. Our results confirm the potential of LLMs for cost-effective labeling of asylum decisions, but highlight the imperfect and inconsistent nature of LLM annotators, and the need to look beyond the predictions of a single, arbitrarily chosen model. The RAB-Cred dataset and code are available at \url{https://github.com/glhr/RAB-Cred}

\end{abstract}

\section{Introduction}

Deepening our understanding of the asylum-decision making process (e.g. discovering bias) requires understanding whether and how the applicant's \textit{credibility} is assessed across a large body of legal decisions. Credibility is a central element in almost all legal proceedings but is known to play an inordinate role in asylum cases, where adjudicators often find evidence to be scarce.

We specifically focus on the Danish asylum decision-making system. When the Danish Immigration Service rejects an asylum application, the case is automatically appealed to the Danish Refugee Appeals Board (RAB) for reassessment. Written decision summaries spanning the past two decades, including the RAB's legal reasoning, are publicly available and form the basis of this work.

When assessing an applicant's eligibility for protection, the RAB often conducts a credibility assessment of the applicant's testimony, that is, whether the applicant's narrative of past events is deemed trustworthy and plausible. To date, credibility assessment in the Danish RAB's decisions has only been studied at a small scale, relying on expert manual annotation~\cite{data-lens-credibility_2022,nordic-asylum-practice_2023,trans-enough-cred_2025}, a time-intensive process. Large Language Models (LLMs), as zero-shot or few-shot text classifiers, offer a potential path to labeling credibility assessments automatically, and thus enabling future large-scale studies in the legal domain.

However, identifying \textit{whether} a credibility assessment was made and whether it was \textit{positive} or \textit{negative} is not straightforward. First, the process of credibility assessment is poorly understood and characterized by fuzziness, with no clear consensus among practitioners and researchers~\cite{credibility-risk-report,jarlner2026credibility}. Thus, the pre-existing general knowledge of off-the-shelf LLMs' must be complemented with a detailed task definition provided by domain experts. Second, even with a specific definition, reliable annotation requires fine-grained semantic and contextual understanding of Danish legal texts.

\begin{table*}[!t]
  \centering
  \begin{tabular}{p{7.5cm}p{7.5cm}}
    \hline
    \multicolumn{2}{p{15.5cm}}{\small{The Board granted a residence permit (Refugee Protection Status) to a female citizen of Somalia, born in 1989. She entered Denmark in February 2004. Like the Immigration Service, \hl{the Board considered the applicant to be a Somali citizen. Based on the information in the case, the Board had to conclude that the applicant had never been to Somalia, did not speak the language, and had no family or other network in Somalia.} As the applicant was a single girl aged 16, the Board found that, based on the background information, it was probable that she would risk inhuman or degrading treatment covered by section 7(2) of the Aliens Act if she were deported to Somalia. The applicant had lived in Yemen since she was five years old without having had any conflicts with the Yemeni authorities. However, it was unclear on what basis she had been residing in Yemen. As it was therefore uncertain whether the applicant had legal residence in Yemen and could enter Yemen, the majority of the Board found that Yemen could not be considered the applicant's first country of asylum pursuant to section 7(3) of the Aliens Act. As a result, the Refugee Appeals Board granted the applicant a residence permit pursuant to section 7(2) of the Aliens Act.
    
    \vspace{\promptspacing}
    
    (\textit{Original case text:} \url{https://fln.dk/praksis/2019/april/somalia-somalia20054/})
}} \\ 
    \hline \hline
    Q1: Credibility assessment present? & Yes (Confidence: Medium)\\
    Q2: Credibility assessment sentiment & Positive (Confidence: High) \\ \hline
    
  \end{tabular}
  \caption{\label{tab:sample_annotation_val}
    Translation of the shortest written decision from the validation set, and its corresponding annotation agreed upon by the two domain experts. Translated with DeepL.com (free version). For reference, we highlight the extract most indicative of a positive credibility assessment for domain experts.
  }
\end{table*}

\textcolor{blue}{In practice, due to the specificities of this data, difficult or atypical cases face a real risk of systematic misclassification by an LLM. Unlike the case outcome or demographic factors which can easily be extracted, annotating credibility assessment might require  ``reading between the lines'', as it is not necessarily stated explicitly (see Table~\ref{tab:sample_annotation_val}). Asylum claims can be rejected despite a positive credibility assessment, or vice-versa, due to the important distinction between credibility assessment (are facts accepted?) and risk assessment (are  there sufficient grounds for asylum, given background material and accepted facts?). These two assessments can be difficult to disentangle, as they are often proximal in articulation and position within the same text; small differences in phrasing can change the final label entirely.  Moreover, a single decision can also contain elements pointing to both positive and negative credibility, when some facts are accepted while others are rejected. These challenges are compounded by the linguistic setting: RAB decisions are written in Danish, a medium-resource language, using specialized legal terminology.}

Domain experts have an interest not only in LLM annotations being as accurate as possible, but also in understanding what \textit{types} of error occur, for \textit{which cases}, and whether these mistakes are \textit{understandable} or \textit{unacceptable}. \textcolor{blue}{Our aim is therefore to investigate the extent to which the annotation of credibility assessment in Danish asylum decisions can be reliably automated by off-the-shelf LLMs, with a particular focus on annotation error.} Our contributions are summarized as follows:

\begin{itemize}[itemsep=0pt,parsep=2pt,topsep=0pt]
    \item we present RAB-Cred, an expertly annotated Danish legal text classification dataset from an under-represented domain and language, which poses interesting challenges for legal experts and natural language understanding. 
    \item to explore the potential of zero-shot and few-shot classification for this task, we systematically benchmark 21 open-weight multilingual LLMs across 30 different prompts.
    \item beyond standard aggregated performance metrics, we analyse the errors produced by and across top-performing models, relating them to human label confidences.
\end{itemize}
The RAB-Cred dataset (including expert multi-annotator labels, and self-reported confidence, and case outcome), along with LLM annotations and code to reproduce the analysis are available at \url{https://github.com/glhr/RAB-Cred}.

\section{Related work}

\textbf{Task and dataset}\quad There is a growing interest in treating LLMs as a drop-in replacement for human annotators/coders in social sciences and humanities \cite{tornberg2024best,ziems-etal-2024-large,davidson2024start,codebook-llms-political-science_2025,wen2025leveraging,llm-annotating-discourse-strategies_2026}. In legal research specifically, off-the-shelf LLMs have been evaluated on a variety of tasks ranging from argument mining~\cite{held-habernal-2025-contemporary} to legal interpretation classification~\cite{dugac2025legal}, showing mixed results. To the best of our knowledge, the identification and sentiment classification of credibility assessment in legal decisions constitutes a novel NLP task, absent from existing legal text classification datasets~\cite{nlp-legal-domain_2025}.

Furthermore, publicly available and annotated texts datasets within refugee law are especially scarce - AsyLex~\cite{asylex_2023} currently being the only one to our knowledge. Unlike AsyLex, RAB-Cred contains non-English texts. Compared to English, Danish is under-represented in LLM training corpora ~\cite{enevoldsen2023danishfoundationmodels,zhang2024snakmodellessonslearnedtraining,ekgren-etal-2024-gpt} and evaluation benchmarks ~\cite{global-mmlu_2025,mega-benchmark_2023,xuan-etal-2025-mmlu}.

\newpage
\textbf{Models and prompts}\quad We include a wide open-weight model selection, including recent multilingual releases such as EuroLLM 22B and Bielik 11B v3~\cite{ramos2026eurollm22btechnicalreport,ociepa2025bielik11bv3multilingual} - this contrasts to the majority of related studies which only consider a small handful of (often proprietary) models~\cite{pavlovic-poesio-2024-effectiveness}. The importance of a well-crafted prompt is repeatedly highlighted in the LLM annotation literature, but the effectiveness of different formulations is largely model, domain and task-specific~\cite{pavlovic-poesio-2024-effectiveness,mizrahi-etal-2024-state}.  We therefore treat prompt choice as a key, model-specific hyper-parameter in our experiments. We benchmark state-of-the-art approaches including chain-of-thought (CoT), metacognitive and few-shot prompting~\cite{vatsal2024survey}, and also experiment with providing varying levels of detail and context, similarly to~\citet{majer-snajder-2024-claim}.

\textbf{Mistakes matter}\quad In order to reflect on the limitations of LLM annotators and on the data itself, we go beyond aggregated classification performance of individual models and zoom into how and when mistakes occur. In a similar spirit,~\citet{codebook-llms-political-science_2025} take a holistic approach when evaluating LLM's ability to follow a codebook for annotation, including manual error analysis and identification of shortcut behaviour. We go a step further and also look at consistency and correctness of LLM-generated annotations \textit{across} models and prompts, at the level of individual samples. Our evaluation draws from error and prompt sensitivity analyses performed in existing work~\cite{majer-snajder-2024-claim,zhuo-etal-2024-prosa}, but we instead consider an ensemble of 15 top-performing LLM annotators. Moreover, input from domain experts enables us to relate LLM misclassifications to  human confidence, and to assess their severity.

\section{The RAB-Cred dataset}

\begin{figure}[h]
    \centering
    \includegraphics[width=0.85\linewidth]{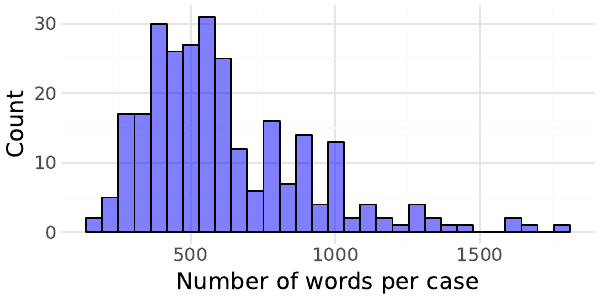}
    \caption{Distribution of case lengths in RAB-Cred.}
    \label{fig:num_words_histogram}
\end{figure}

The dataset used in this work is based on public asylum decisions made by the Danish Refugee Appeals Board (RAB), available at \url{https://fln.dk/praksis}. The RAB is the second and final legal instance of the Danish asylum system. Cases  rejected by the Immigration Service are automatically appealed to the Board. Written decisions are relatively brief (around 600 words on average) - although some extend beyond 1500 words, cf. Figure~\ref{fig:num_words_histogram}. They first outline details about the case and asylum motives, and then explain the RAB's decision to either uphold or over-turn the Immigration Service's rejection, or to remand the case. Additional information about the data, metadata and annotations can be found in Appendix~\ref{app:dataset}.

Given a written decision, the aim is to identify \textit{whether a credibility assessment was made}, and if so, whether credibility was assessed \textit{positively or negatively}.

\subsection{Validation and test sets}

We sample 273 cases from the RAB's official website using stratified random sampling across time ranges of interest (cf. Appendix~\ref{app:dataset} for details). 73 of these cases are used as a \textit{validation} set, for the human annotators to iteratively develop a codebook and for selecting optimal model-prompt combinations when generating LLM annotations. The remaining 200 cases are held out as an \textit{unseen test} set, which we use to quantify human inter-annotator agreement and analyse the errors made by LLMs.

\subsection{Codebook and class definition}

Unlike \citet{llm-thematic-analysis-legal-studies_2023,mistral-paper-sabet} who involve LLMs in designing the annotation task itself (e.g. identifying relevant concepts to annotate), we rely on a codebook developed by refugee law experts, which defines the categories and annotation guidelines. This codebook is used as a basis for both human annotation and LLM prompts - an approach which has shown promising results in recent work \cite{ruckdeschel-2025-just,codebook-llms-political-science_2025}.

Specifically, two domain experts and annotators (nicknamed H1 and H2) set out to jointly define what exactly constitutes a credibility assessment in the RAB dataset, and how it should be annotated at the text level. Through discussions and interactive annotation sessions, the codebook was iteratively refined until full inter-annotator agreement was reached on the validation set. The annotators converged to a 3-tiered categorization, where the credibility assessment is annotated as either Absent, Positive, or Negative. For each case, the annotators also recorded their confidence level (Low, Medium or High) about the presence of a credibility assessment, and its sentiment (if present). Note that these are posed as two separate questions, as an annotator may be highly uncertain about whether certain statements qualify as a credibility assessment, but may be highly confident that \textit{if} they qualify as a credibility assessment, the assessment is positive. Table~\ref{tab:sample_annotation_val} shows an example annotation.

\begin{figure}[h]
    \centering
    \includegraphics[width=\linewidth]{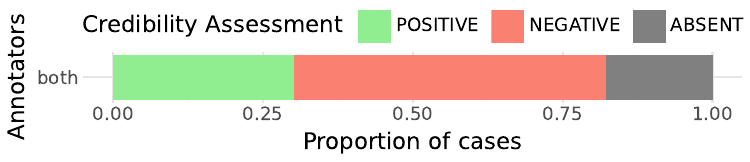}
    
    \vspace{0.5em}
    
    \includegraphics[width=\linewidth]{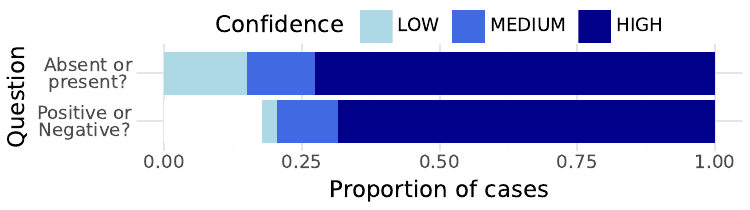}
    \caption{Label (top) and confidence (bottom) distribution of human annotators on the validation set.}
    \label{fig:val_set_distribution}
\end{figure}

Figure~\ref{fig:val_set_distribution} shows that according to the two experts, a credibility assessment is present in over 75\% of cases; when present, it is more likely to be negative. Looking at annotator confidence suggests that while the experts are largely confident in their annotation, the \textit{presence} of a credibility assessment is more difficult to ascertain than its sentiment.

\subsection{Inter-annotator agreement on the test set}

\begin{figure}[t]
    %\centering
    %\begin{subfigure}{\linewidth}
    \centering
    \includegraphics[height=3.9cm]{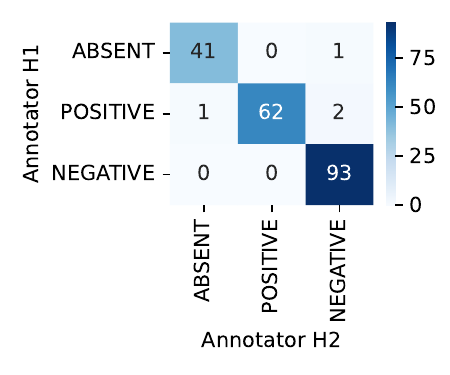}
    \caption{Confusion matrix showing inter-annotator agreement on the test set (H1 vs. H2's annotations).}
    \label{fig:confusion_matrix_annotators}
    %\end{subfigure}
\end{figure}

Gold-standard annotation of the test set was performed by the same two domain experts H1 and H2 as for the validation set, but completely independently. We observed a very high level of agreement on the presence and sentiment of a credibility assessment (Cohen's $\kappa=0.97$), with only 4 cases (out of 200) for which the annotators diverge - as shown in Figure~\ref{fig:confusion_matrix_annotators}. Interestingly, for 3 out of these 4 cases, both annotators reported high confidence. The high inter-rater agreement suggests that the annotations are of high quality and that the codebook is sufficiently informative, but the presence of high-confidence disagreement suggests that even with a strong, detailed understanding of the codebook, some ambiguity or possible conflicting interpretation remains.

\begin{figure}[t]
    \centering
    \includegraphics[width=\linewidth]{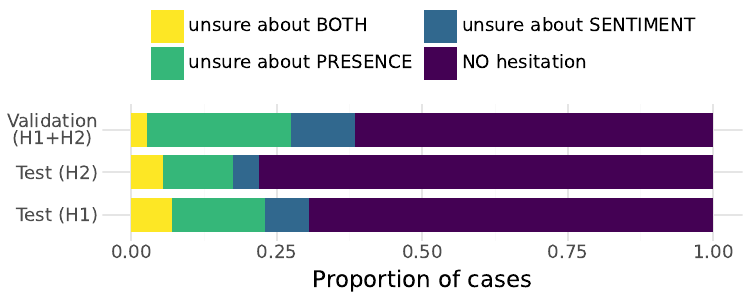}
    \caption{Inter-class hesitation of the human annotators, inferred from annotator confidence.}
    \label{fig:human-hesitation}
\end{figure}

As shown in Figure~\ref{fig:human-hesitation}, when independently annotating test samples, both annotators are more likely to be unsure about the presence of a credibility assessment than its sentiment.

To resolve disagreement between H1 and H2, a third domain expert H3 (who was not involved in codebook development and was not shown H1 and H2's labels) annotated the 4 cases; for all 4, H3 aligned with either H1 or H2. The final gold label is taken as the majority vote.

    %\vspace{1em}

\subsection{Correlation with outcome}

\begin{figure}[h]
    %\begin{subfigure}{\linewidth}
    \centering
        \includegraphics[height=3.3cm]{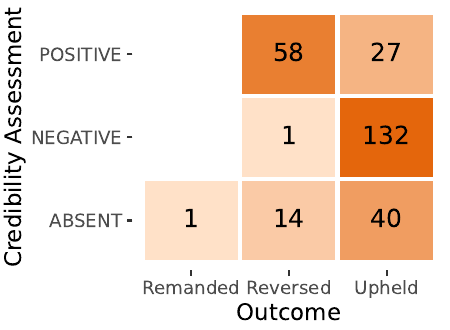}
        \caption{Relation between case outcome and gold credibility assessment labels on the full dataset.}
        \label{fig:confusion_matrix_outcome}
    
\end{figure}

Figure~\ref{fig:confusion_matrix_outcome} shows a notable relation between case outcome and credibility: 80.6\% of reversed cases in RAB-Cred (asylum granted) are associated with a positive credibility assessment, and 66.3\% of upheld rejections with a negative one. At the same time, for 41.5\% of cases, either no credibility assessment is present or the outcome contradicts the credibility sentiment. The outcome should therefore not be used as a proxy for annotation.

\section{LLM-generated annotations}

\textcolor{orange}{Given the small amount of human-labeled data at our disposal, our annotation pipeline relies on off-the-shelf LLMs used as zero-shot or few-shot classifiers. We first describe our selection of language models, along with the prompt variants used to evaluate automatic credibility annotations. We then systematically evaluate the role of model choice and prompt choice for this task. The full list of models and implementation details are described in Appendix~\ref{app:model_selection}, the prompts and the few-shot examples are in Appendix~\ref{app:prompt-variants}, and detailed results in Appendix~\ref{app:full-perf-val}.}

\subsection{Models and configuration}

When selecting potential off-the-shelf LLMs, we considered a wide range of instruction-tuned models and applied the following criteria: 

% [itemsep=0pt,parsep=2pt,topsep=2pt]
\begin{enumerate}[itemsep=0pt,parsep=2pt,topsep=2pt]
    \item \textbf{Open-weight} to ensure reproducibility and also to enable offline annotation, which is necessary for annotating sensitive data.
    \item Explicitly trained on \textbf{multilingual} data.
    \item \textbf{Limited model size} due to compute constraints (a single H100 with 80 GB VRAM).
    \item \textbf{Context length} of at least 8K tokens, to accommodate long case texts, detailed instructions and LLM reasoning output.
\end{enumerate}
This resulted in 21 models spanning 9 model families (Gemma~\cite{gemmateam2025gemma3technicalreport}, Qwen~\cite{qwen2.5}, Phi~\cite{abdin2024phi4technicalreport}, Aya~\cite{dang2024ayaexpansecombiningresearch}, Granite~\cite{granite4}, Mistral~\cite{liu2026ministral3}, Llama~\cite{llama_2024}, Bielik~\cite{Bielik11Bv3i-model-card} and EuroLLM~\cite{ramos2026eurollm22btechnicalreport}), ranging from 3B to 35B parameters.

\subsection{Prompt variants}

We separately consider the role of the \textit{system} prompt and the \textit{user} prompt for annotating credibility. The system prompt is used to provide background knowledge (e.g. what is a credibility assessment?), while the user prompt determines how the classification task should be tackled (e.g. directly giving the final answer, using examples or following multiple steps).

We design 6 system prompts and 5 user prompts, yielding a total of 30 unique prompts templates. All prompts are written in English with Danish case text as input, as~\citet{lai-etal-2023-chatgpt,pavlovic-poesio-2024-effectiveness} found this approach effective for non-English data.

\paragraph{System prompts (SP)}\quad System prompts (SP) follow a nested hierarchy of increasing domain context. SP0 provides no system prompt (baseline). SP1 assigns a domain-expert persona. SP2 extends SP1 with the verbatim codebook. SP3 restructures SP2 with indicative Danish phrases per class. SP4 extends SP3 with critical edge cases (e.g., hypothetical legal constructions, mixed sentiment). SP5 extends SP1 by offering an alternative expert-written breakdown that explicitly disambiguates credibility from risk assessment and neutral reporting.

\paragraph{User prompts (UP)}\quad \textcolor{blue}{We design 5 user prompts of increasing complexity. \hyperref[box:UP1]{UP1} directly instructs the model to select one of three classes. \hyperref[box:UP1-FS]{UP1-FS} extends \hyperref[box:UP1]{UP1} with three labeled examples (selection described below). \hyperref[box:UP2]{UP2} decomposes the task into two binary questions: first whether a credibility assessment is present, then its sentiment, mirroring the human annotation structure. \hyperref[box:UP1-FS]{UP3} and \hyperref[box:UP1-FS]{UP4} introduce an unconstrained reasoning step before the final classification: \hyperref[box:UP1-FS]{UP3} via zero-shot CoT prompting ~\cite{kojima2022large} and \hyperref[box:UP1-FS]{UP4} via zero-shot metacognitive prompting ~\cite{wang2024metacognitive}.}

\paragraph{Few-shot examples}\quad We select one example per class from the validation set (thus three examples in total), choosing cases that are unambiguous for domain experts (high label confidence) yet consistently challenging for LLMs (highest zero-shot misclassification rate in preliminary experiments).

From a domain perspective, each example represents an atypical scenario: the "absent" case lacks a credibility assessment; the "positive" and "negative" cases each exhibit a mismatch between credibility sentiment and outcome.

\subsection{Comparing models and prompts}\label{sec:val-comparison}

We first investigate which models and prompting strategies are best suited for annotating credibility assessment. We evaluate the 21 selected models $\times$ 30 user-system prompt combinations on the validation set - excluding the 3 few-shot samples used in UP1-FS, leaving 70 samples.

Given class imbalance, we report macro F1 score as the primary metric for classification performance. As a lower baseline, the outcome is used as a naive heuristic for credibility assessment using the following mapping: remanded $\rightarrow$ absent, reversed $\rightarrow$ positive, upheld $\rightarrow$ negative.

\begin{figure*}[!t]
    \centering
    \includegraphics[width=\linewidth]{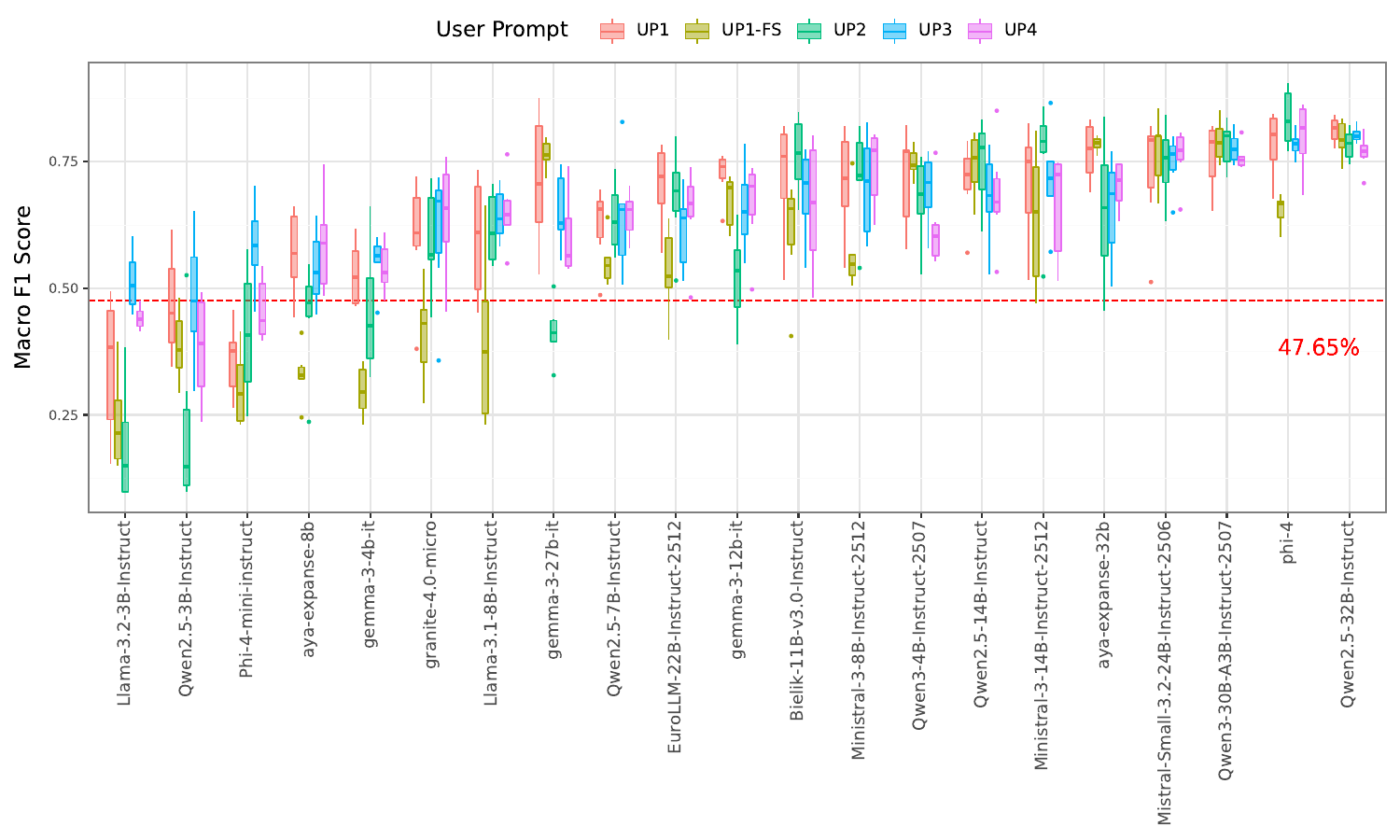}
    \caption{Validation set classification performance per model for different user prompts. Each boxplot is across 6 system prompts. Models are ordered by average macro-F1 score. Red line: outcome-as-credibility baseline.}
    \label{fig:f1_scores_by_model_and_prompt_variant.pdf}
\end{figure*}

 \begin{figure}[!t]
    \centering
    \includegraphics[width=\linewidth]{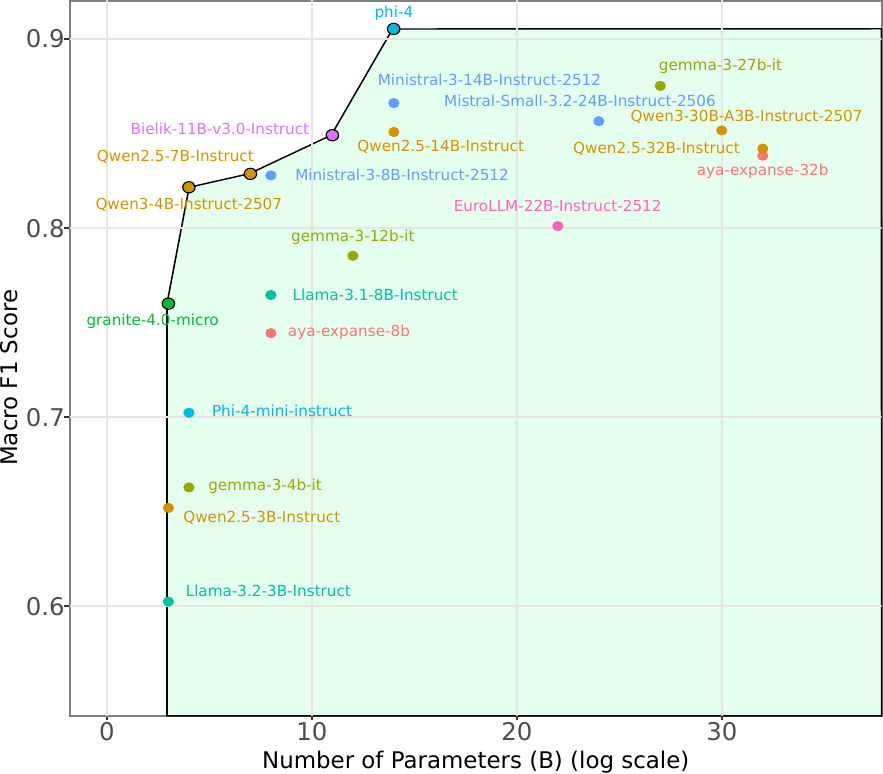}
    \caption{Best-case classification performance (taking the top 1 system-user prompt for each model) on the validation set, as a function of model size. }
    \label{fig:best-f1-modelsize}
\end{figure}

\paragraph{Comparing models}

Figure~\ref{fig:best-f1-modelsize} shows the size-performance tradeoff considering only the best system-user prompt per model, while Figure~\ref{fig:f1_scores_by_model_and_prompt_variant.pdf} shows model-specific performance across all prompt variants. With the exception of Qwen2.5-7B and Ministral-3-8B, all models that consistently outperform the outcome-as-credibility baseline are larger than 11B. However, as shown in Figure~\ref{fig:best-f1-modelsize}, increases in size do not necessarily translate to performance gains. Among models under 10B, Qwen2.5-7B and Qwen3-4B offer the best trade-off between size and performance. Perhaps most strikingly, phi-4 (14B) achieves the highest F1 on the validation set (90.51\%), while the largest models in our selection, including Qwen2.5-32B, Qwen3-30B and aya-expanse-32b, plateau at or below 85\% F1.

\begin{figure}[!h]
    \centering
    \includegraphics[width=0.9\linewidth]{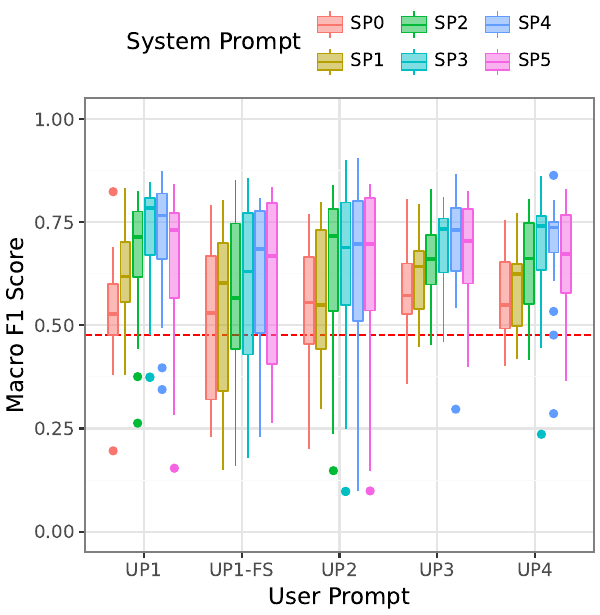}
    \caption{Validation set classification performance across 21 models for different user-system prompt combinations. Red line: outcome-as-credibility baseline.}
    \label{fig:f1_scores_by_user_prompt_and_system_prompt}
\end{figure}

As shown in Figure~\ref{fig:f1_scores_by_user_prompt_and_system_prompt}, models vary considerably in how sensitive they are to different prompt combinations. Models such as Qwen2.5-32B are highly robust, while performance for a model like Gemma-3-27B can vary by over 54\% in F1. A general tendency is that bigger models vary less in performance with prompt changes. In the absence of any explanation of what a credibility assessment is (SP0 or SP1), the 2 largest Qwen models and phi-4 stand out as the best performing models, with Qwen2.5-32B exceeding 83\% F1 using UP1 both with (SP1) and without (SP0) a persona.

\paragraph{Comparing system prompts}

Looking at Figure~\ref{fig:f1_scores_by_user_prompt_and_system_prompt}, although the relative effectiveness of different SPs is UP-dependent, there is a clear benefit in incorporating expert knowledge (SPs 2-5), with SP3 and SP4 showing the most consistent performance. This shows the benefit of interdisciplinary prompt design:  different from SP2 and SP5 which were written by domain experts alone, SP3 and SP4 were formulated by computer scientists based on the codebook and with domain expert feedback. Adding SP4's edge-cases to SP3's handcrafted context is effective in some cases, but not on average.

\paragraph{Comparing user prompts}

Compared to the basic UP1 prompt, the effect of more advanced prompting strategies (few-shot, multi-turn, and CoT/metacognitive) is found to be highly model-specific. For instance, while Phi-4 greatly benefits from having the classification task broken down into 2 binary questions with UP2 (such that it achieves the highest performance of any model-prompt combination), this approach heavily degrades performance for Gemma models. We find few-shot prompting (UP1-FS) to be effective only for Qwen3-30B, Mistral-Small and Qwen2.5-4B. As for reasoning-based prompts (UP3 and UP4), these reduce the variation in performance across SPs and models (Figure~\ref{fig:f1_scores_by_user_prompt_and_system_prompt}). CoT prompting only outperforms UP1 when the SP lacks sufficient context (SP0 and SP1.)

\paragraph{Summary} Overall, while adding explanations of credibility assessment through the system prompt clearly helps, no single prompt emerges as the ``winner'', which confirms the need for model-specific prompt choice. In the zero-shot setting, Phi-4 shows impressive performance for its size, see Figure~\ref{fig:best-f1-modelsize}. This is somewhat surprising, given that it is seldom evaluated in the LLM-as-annotator literature, and its model card states ``multilingual data constitutes about 8\% of our overall data'' and ``phi-4 is not intended to support multilingual use''.

\section{When and how do the best models fail?}

\textcolor{orange}{Our aim is to analyze classification errors and agreement among top-performing LLM annotators. We select the most promising model-prompt combinations based on performance on the validation set. Since the optimal prompt is highly model-dependent, we rank each model according to its average macro-F1 score across its top-3 prompts, then select the top 5 models (phi-4, gemma-3-27b-it, Ministral-3-14B-Instruct-2512, Ministral-3-14B-Instruct-2512, Qwen3-30B-A3B-Instruct-2507 - each paired with its 3 highest-performing prompts) for final evaluation on the test set. The resulting selection contains 11 unique user-prompt combinations, with SP5+UP2 being the most frequent (3 instances). Individual classification performance for these 15 model-prompt combinations is reported in Appendix~\ref{app:full-perf-val} and~\ref{app:selected-perf-test}.}

\textcolor{orange}{When compared to the majority label assigned by human annotators, the macro F1-Score of the selected model-prompt combinations ranges from 84.4\% to 94.7\% on the test set, with phi-4 being the strongest model, and gemma-3-27b-it being the weakest on average. For reference, the outcome-as-classifier baseline achieves an F1-Score of 53\%.}

\begin{table}[h]
    \centering
    \resizebox{\linewidth}{!}{%
    \begin{tabular}{rrcc}
        \toprule
        \multicolumn{2}{r}{\textbf{annotator pair}} & \makecell{\textbf{agreement with}\\\textbf{human majority}} & \makecell{\textbf{inter-annotator}\\\textbf{agreement}} \\ \midrule
       \multirow{2}{*}{\footnotesize{Domain experts}} & H1 & 0.984 & \multirow{2}{*}{0.967} \\
        & H2 & 0.983 &  \\
        \midrule
        \multirow{2}{*}{\footnotesize{Mistral-Small-3.2-24B-Instruct-2506}} & SP3+UP1-FS & 0.883 & \multirow{2}{*}{0.922} \\
        & SP4+UP1 & 0.882 & \\ \midrule
        \multirow{2}{*}{\footnotesize{phi-4}} & SP4+UP2 & 0.906 & \multirow{2}{*}{0.913} \\
        & SP4+UP4 & 0.913 & \\
        \bottomrule
    \end{tabular}%
    }
    \caption{Cohen's $\kappa$ for 3 annotator pairs on the test set.}
    \label{tab:agreement-human-llm-pairs}
\end{table}

\vspace{-1em}

\paragraph{Inter-LLM agreement}

\begin{figure*}[!t]
    \centering
    \includegraphics[width=\linewidth]{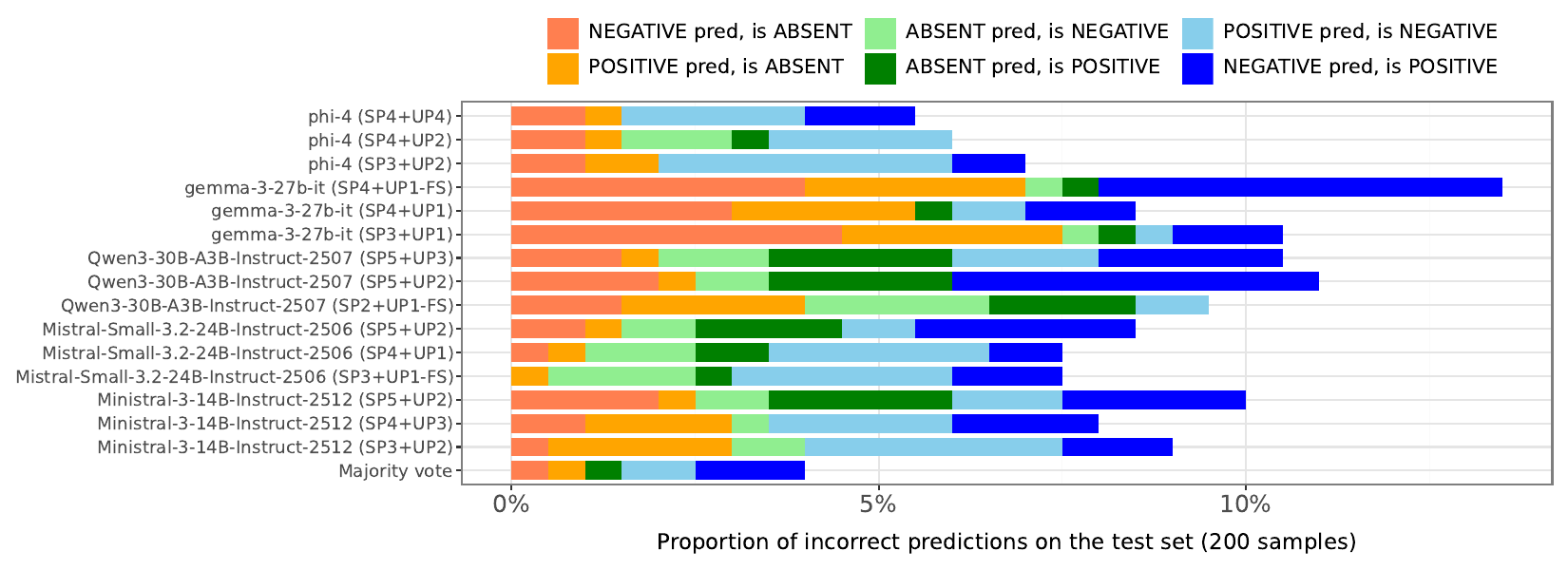}
    \caption{Individual LLM and ensemble mistakes, color-coded by class confusion (cf. Appendix~\ref{app:test-interclass-confusion}).}
    \label{fig:credibility_pred_classes}
\end{figure*}

We compare the human inter-annotator agreement on the test set (between H1 and H2) with that of LLM annotators in Table~\ref{tab:agreement-human-llm-pairs}. We select two pairs of LLM annotators with near-identical agreement levels with respect to the human majority label (i.e. similar level of ``correctness''), and measure the agreement within the pair. We find the LLM annotator pairs to be less aligned than the human annotator pair, even when they share the same model weights and system prompt, indicating that misclassifications are not necessarily consistent across similarly-performing models. The inter-annotator agreement for all possible LLM annotator pairs can be found in Appendix~\ref{app:model-agreement-test}. The highest level of agreement ($\kappa = 0.950$) occurs between two Gemma models with the same user prompt, but different system prompts (SP3 vs. SP4), despite their F1 differing by almost 3\%.

\paragraph{Instance-level sensitivity}

\textcolor{blue}{Prompt sensitivity is often measured at the dataset level, looking at how aggregate performance measures vary across different prompts~\cite{hua-etal-2025-flaw,mizrahi-etal-2024-state}, similarly to~Section \ref{sec:val-comparison}. However, two LLM annotators with the same performance may be misclassifying different instances. To complement our aggregate analysis, we adopt the PromptSensiScore (PSS) from~\citet{zhuo-etal-2024-prosa}, which instead operates at the instance level:  it captures changes in the correctness of individual predictions, given a change of prompt. To compare the effect of model vs. prompt choice, we apply PSS in two ways: fixing the model and varying the prompt vs. fixing the prompt and varying the model (detailed results in Appendix~\ref{app:sensitivity-test}).}

\textcolor{blue}{Examining prompt sensitivity first, we find that phi-4, gemma-3-27b, and Mistral-Small-24B all exhibit lower instance-level sensitivity than Ministral and Qwen3. Interestingly, Qwen3-30B shows the highest sensitivity despite strong aggregate stability (F1 score variation across prompts), suggesting its consistent performance masks instability on specific instances. This pattern underscores that dataset-level and instance-level metrics capture complementary axes of robustness.}

\textcolor{blue}{For model sensitivity, we select a fixed prompt (SP5+UP2) shared across Ministral-14B, Mistral-Small-24B, and Qwen3-30B, and look at instability related to model change. The resulting PSS score of 0.05 is lower than model-specific PSS scores (Qwen: PSS$\approx$0.10, Ministral: PSS$\approx$0.08, Mistral: PSS$\approx0.06$), indicating that prompt variations introduce greater instability than model differences, and underscoring the importance of prompt design.}

\paragraph{Inter-class confusion} In Figure \ref{fig:credibility_pred_classes}, we zoom into the number and types of mistakes made by the 15 individual LLM annotators and by the ensemble (taking the majority vote). For individual LLMs, we find that the relative prevalence of different types of mistakes varies significantly across both models and prompts. Some LLMs never miss the presence of a credibility assessment (green in Figure~\ref{fig:credibility_pred_classes}), however all LLMs falsely identify a credibility assessment at least once (orange/salmon). Distinguishing between positive vs. negative credibility assessments seems less straightforward for LLM annotators than for human annotators (cf. Figure~\ref{fig:human-hesitation}): when taking the majority vote, over half of the mistakes are \textit{sentiment} misclassifications.

\begin{figure*}[!h]
    \centering
    \includegraphics[width=\linewidth,trim={0 20 0 0},clip]{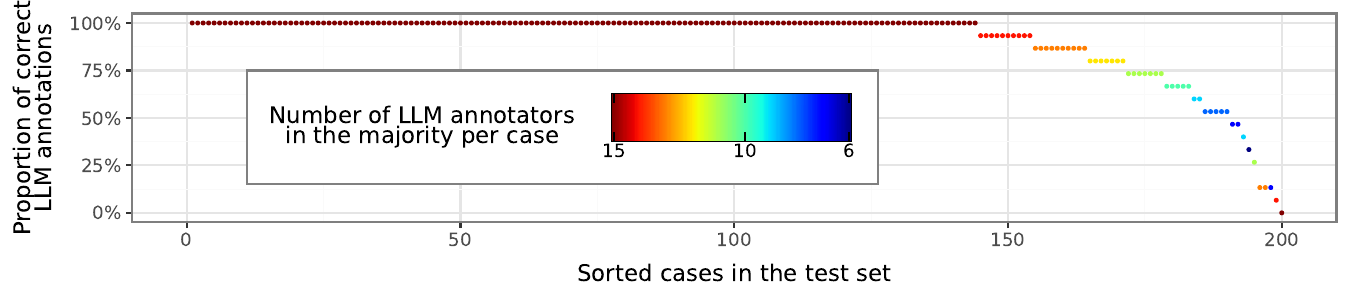}
    \includegraphics[width=\linewidth]{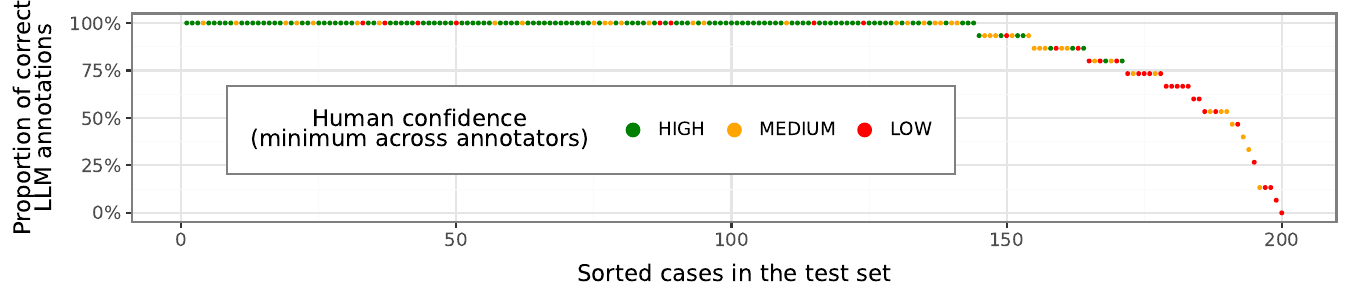}
    \caption{LLM agreement  vs. LLM correctness vs. human confidence. Each point corresponds to a single case in the test set, and the points in both plots are sorted by correctness (identical ranking in both plots).}
    \label{fig:test_casewise_accuracy_human_conf}
\end{figure*}

\paragraph{Fine-grained analysis} The small number of test set samples allows us to visualize the correctness and agreement of LLMs on a case-by-case basis, as shown in Figure~\ref{fig:test_casewise_accuracy_human_conf}. 72\% of cases (144 out of 200), are correctly classified by all 15 LLM annotators, 95\% (190 out of 200) are correctly classified by at least half (i.e. at least 8 LLM annotators out of 15). Interestingly all the cases correctly classified by less than 75\% of LLM annotators were assigned medium or low confidence by one of the annotators.

When taking the majority vote across LLM annotators, 96\% of cases are correctly classified - 1.5 percentage points above the top single-model accuracy. For the remaining 8 misclassified cases (cf. Appendix~\ref{app:test-interclass-confusion}), we asked domain expert H1 to judge the LLM majority prediction: H1 considers the misclassification "acceptable" in 4 cases, "understandable" in 2 cases and "unacceptable" in 2 cases. Further details about this categorization are in Appendix~\ref{app:misclassified-cases-test}. We find that \textit{all} model-prompt combinations make at least 1 unacceptable mistake.

Lastly we zoom into the two cases which were misclassified by 14 or all 15 LLM annotators respectively (bottom right of the plots in Figure~\ref{fig:test_casewise_accuracy_human_conf}): both were labeled as having no credibility assessment by H1 and H2. Interestingly, H1 considers the LLM majority prediction to be acceptable in both cases, to the point of reconsidering their own annotation. In one case, hesitation is due to the use of language typically associated with credibility assessments but a future-oriented judgment, while in the second case, credibility assessment of the claimant's relative may be misconstrued with the claimant's own credibility.  We refer to Appendix~\ref{app:misclassified-cases-test} for the two case texts, LLM reasoning and domain expert reasoning.

\section{Discussion and future work}

From a practical standpoint, our results serve as a solid baseline for the RAB-Cred dataset and suggest that automating the annotation of credibility assessment using LLMs is a promising direction, but not a perfect replacement for manual expert annotation. The case-by-case analysis provides preliminary support for the use of prompt and model ensembling, as LLM aggregation yields a clear improvement over any single LLM annotator. Furthermore, ensembling of LLM annotators could enable a human-in-the-loop approach where cases with high inter-model or inter-prompt disagreement are flagged for expert review, while the rest are annotated automatically.  A systematic comparison of ensembling strategies and their cost-performance trade-off is a natural direction for future work.

\newpage
A consistent finding across our experiments is that prompt design matters at least as much as model choice. The effectiveness of advanced prompting strategies is highly model-dependent, and at the instance level, changing the prompt can affect individual predictions more than changing the model. This confirms the need for multi-prompt evaluation in LLM annotation studies~\cite{mizrahi-etal-2024-state}. It also suggests that multi-prompt ensembling with a single model could be sufficient.

Furthermore, several design choices in our study point to avenues for further investigation. We use English-language prompts on Danish input texts, following evidence that this cross-lingual approach is effective for multilingual classification~\cite{lai-etal-2023-chatgpt, pavlovic-poesio-2024-effectiveness}. However, the interaction between prompt language, input language, and model choice is underexplored for legal texts specifically; prompting in Danish or translating case texts to English may yield different error profiles. Similarly, we evaluate only off-the-shelf, general-purpose LLMs. Domain-specialized or fine-tuned models, as advocated by \citet{dominguez-olmedo2025lawma}, may achieve higher accuracy, particularly on the boundary cases identified in our error analysis—though this comes at the cost of requiring labeled training data, which our approach aims to circumvent. While we focus on the final classification output, systematically analysing the intermediate reasoning traces produced by chain-of-thought and metacognitive prompts could offer further insight into \emph{how} models arrive at their classification and whether this aligns with expert-provided explanations.

\clearpage
\section*{Limitations}

Our evaluation is based on a dataset of 273 cases, of which 3 are used as few-shot examples, 70 as a validation set, and 200 as a test set. While carefully annotated by experts, this dataset is not representative of the full body of publicly available Danish RAB decisions, which spans over 10,000 cases. The relatively small size limits the statistical power of comparisons between model-prompt combinations and may not capture the full diversity of credibility assessment formulations found in practice.

In terms of model selection, we restrict our evaluation to open-weight models of at most 35B parameters, due to compute constraints and the practical need for offline inference when working with sensitive legal data. We do not evaluate closed-source models such as GPT-4o or Claude, which may achieve stronger performance but cannot be deployed locally. Our findings therefore characterize the current capabilities of open-weight, moderately-sized LLMs, and should not be taken as an upper bound on what LLMs can achieve for this task.

Furthermore, all results are based on a single run per model-prompt combination. While we use greedy decoding throughout, which is deterministic for a given input, we do not account for possible stochasticity arising from numerical precision or hardware differences.

Whenever possible, we used constrained decoding via the \textit{outlines} library to ensure that models produce valid category labels (cf. Appendix~\ref{app:constrained-gen} for details). Despite its practical advantages, constrained decoding is known to potentially degrade generation quality~\cite{hidden-cost-constrained-2025}, and its interaction with classification performance across different model architectures has not been systematically studied here.

Finally, although our stratified sampling partially addresses temporal variation in RAB decisions, we do not analyse whether model performance varies across time periods. Changes in legal practice, writing conventions, or anonymization procedures over the two decades covered by the dataset may introduce systematic differences that have not been captured in the evaluation.

\newpage
\section*{Acknowledgments}

This work was supported by the Villum Foundation (“XAI-CRED”, grant no. 69198), the Grundfos Foundation (“REPAI”, grant no. 83648813), and the Danish National Research Foundation ("Center of Excellence for Global Mobility Law", grant no. DNRF169).

Part of the computation done for this project was performed on the UCloud interactive HPC system, which is managed by the eScience Center at the University of Southern Denmark. Part of the computation was also performed on the AI Cloud HPC system managed by CLAAUDIA at Aalborg University.


\bibliography{custom}

\clearpage

\appendix
\renewcommand\thefigure{\thesection.\arabic{figure}}    
\renewcommand\thetable{\thesection.\arabic{table}}    
\setcounter{figure}{0}    
\setcounter{table}{0}    

\onecolumn

\section{Dataset details}\label{app:dataset}

Here we present details about how the RAB-Cred dataset was sampled and annotated, and how we extract the outcome for each case.

\subsection{Intended use}

We provide multi-annotator labels and confidence levels for RAB written decisions indicating the presence and sentiment of credibility assessment, as defined by domain experts, along with case metadata (e.g. outcome). These annotations and metadata must only be used for research purposes, and are solely intended to be used for understanding \textit{whether} and \textit{how} credibility assessment is performed in asylum decisions. They are \textbf{not} intended to be used for assessing/classifying the credibility or veracity of claims made by the applicant. Furthermore, they are \textbf{not} intended to be used for automating decisions.

\subsection{Dataset source}

The dataset was collected via web-scraping  of all written decisions available at \url{https://fln.dk/praksis/} in June 2025 and early December 2025. Combined, this yielded 10817 unique cases dating from 2004 to 2025, from which we sampled two subsets to annotate (val and test).

\paragraph{Representativeness}\quad Not all RAB decisions are published on the website. The website states:  "The Refugee Appeals Board's website regularly publishes summaries of selected decisions that represent the Board's practice regarding individual countries. This means that not all of the Board's decisions are published on the website."\footnote{"På Flygtningenævnets hjemmeside offentliggøres løbende praksisresumeer af udvalgte afgørelser, der udgør et repræsentativt udsnit af nævnets praksis vedrørende de enkelte lande. Det er således ikke alle nævnets afgørelser, der offentliggøres på hjemmeside." \url{https://web.archive.org/web/20260302132511/https://fln.dk/information_til/advokater/naevnets_praksis/}} (translated with DeepL.com). The RAB's specific selection criteria is unknown.

Empirically, we compared the recognition rate (outcome) and country of origin distribution of the scraped data to the yearly statistics published by the RAB\footnote{\url{https://fln.dk/statistik_og_maaltal/}}. Overall, before 2015 we have a slight over-representation of over-turned cases. In terms of country of origin, the scraped data also resembles the top-3 national distribution reported by the RAB, but with a slight under-representation of soviet states and over-representation of Middle East cases.

\paragraph{Anonymization} Written decisions published by the RAB are pseudo-anonymized. Although the specific pseudo-anonymization procedure and criteria is unknown and appears to have evolved over time, the following is stated on the RAB's website: "The practice summaries reproduce the Refugee Board's reasoning in each individual decision in full. However, in some cases, names, dates, locations, etc. have been anonymised for the sake of the applicant." (translated with DeepL.com)]\footnote{"I praksisresumeerne gengives Flygtningenævnets præmis i den enkelte afgørelse i sin fulde længde. Der er dog i nogle tilfælde af hensyn til ansøgeren foretaget en anonymisering af navne, tidsangivelser, stedsangivelser etc." \url{https://web.archive.org/web/20260302132511/https://fln.dk/information_til/advokater/naevnets_praksis/}}.

In practice, we observe that no names or initials of applicants or their relatives are present in any case texts. Furthermore, in cases from 2010 or later, specific details (e.g. a date, age, country, city, ethnicity, medical issue, social media platform, among others) are redacted in the written decisions, and replaced by square brackets. 

\subsection{Sampling of the validation and test sets}

\begin{figure}[h]
        \includegraphics[width=0.47\textwidth]{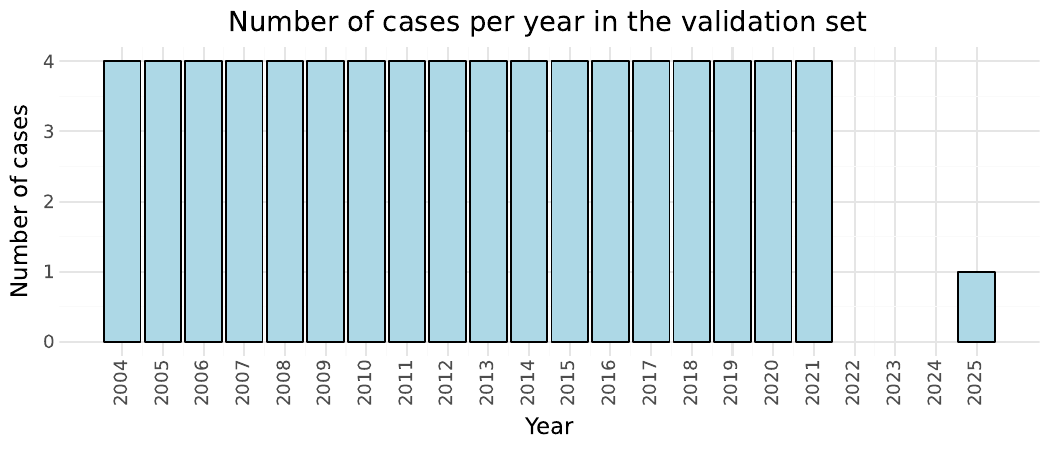}
        \hfill
        \includegraphics[width=0.47\textwidth]{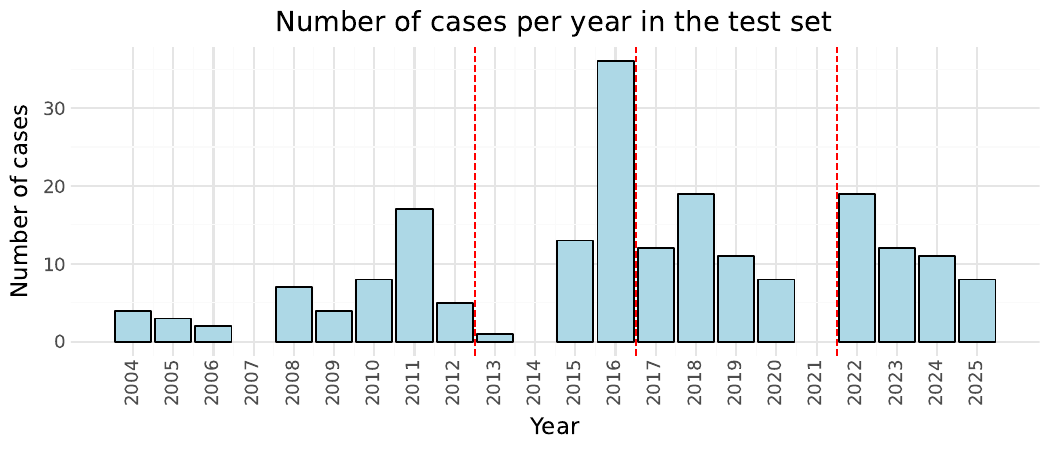}
        \caption{Number of cases in the RAB-Cred dataset by year. The red lines show the 4 time ranges used for stratified sampling of the test set.}
        \label{app:year-distribution}
\end{figure}

The validation set and test set were sampled separately from the 10817 web-scraped cases. Sampling was performed by the domain experts H1 and H2. Yearly distribution for the validation and test sets can be seen in Figure~\ref{app:year-distribution} and is detailed below.

\paragraph{Validation set} Cases were randomly sampled with yearly stratification across the year range 2004 to 2021 (18 years), with 4 cases per year.  In addition, 1 recent case (from 2025) was added due to its difficult nature (surplace case, multiple asylum motives). This results in 73 cases.

\paragraph{Test set} Cases cover a range of years between 2004 to 2025, where time ranges for stratification are based on changes to the board: 2004-2012, 2013-2016, 2017-2021, 2022-2025. The board changes is in its size and composition, and thus could be affected by slight changes in writing style and possibly practice.

\subsection{Human annotators}

Annotation of the validation and test set was performed by the same two annotators (H1 and H2). A third annotator (H3) was introduced to resolve the 4 test set samples where H1 and H2 assigned conflicting labels.

All three annotators are fluent Danish speakers (H2 and H3 being native speakers), with a background in social science and several years of research experience in the field of Danish refugee law, and highly familiar with both the context and the content of the RAB decision texts.

\subsection{Test set annotations}

\begin{figure}[h]
    \centering
    \includegraphics[width=0.7\linewidth]{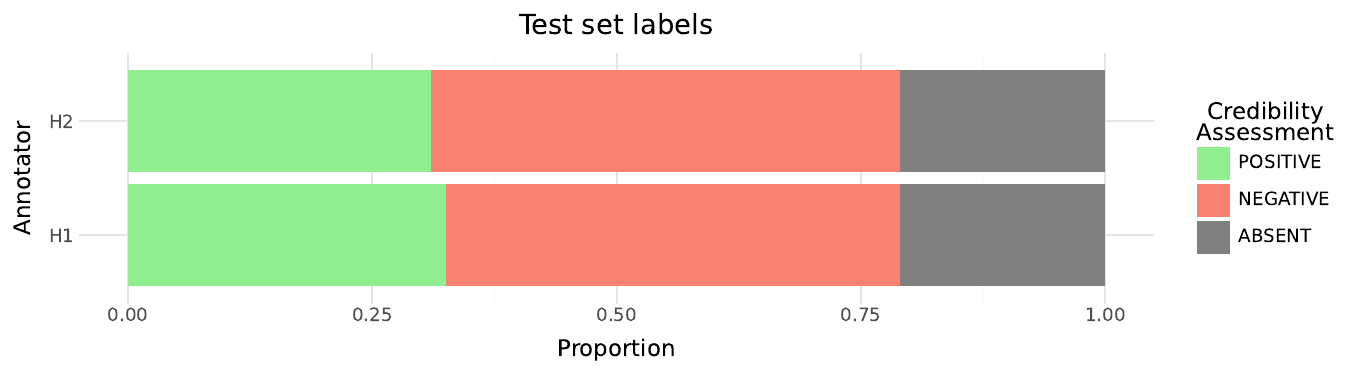}
    
    \vspace{1em}
    
    \includegraphics[width=0.7\linewidth]{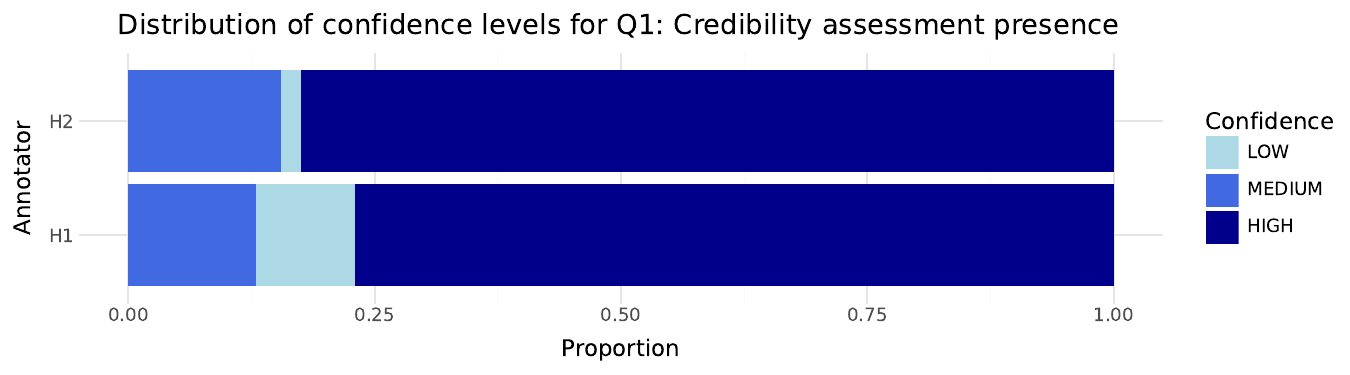}

    \vspace{0.5em}
    
    \includegraphics[width=0.7\linewidth]{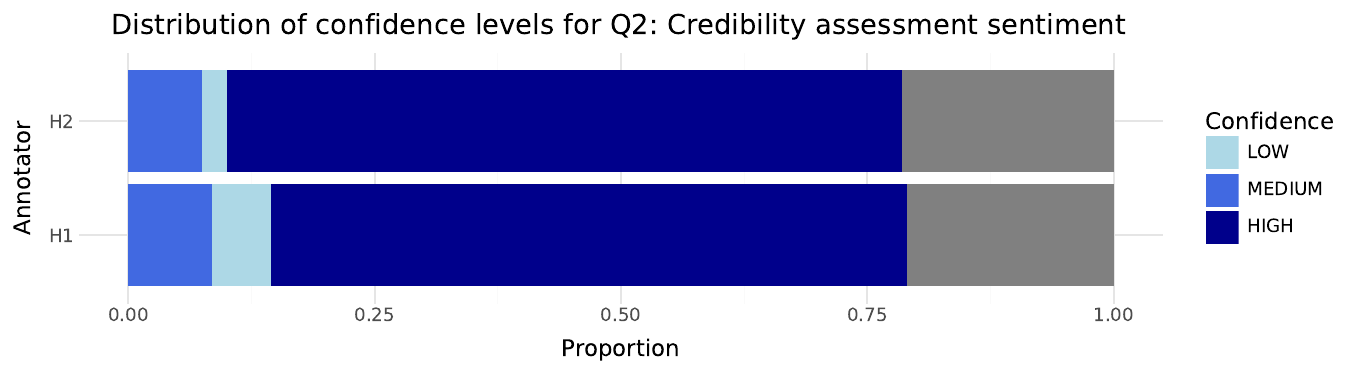}
    \caption{Distribution of labels independently assigned by the two domain experts on the test set (top) and their confidence level for the presence and (optionally) sentiment of a credibility assessment.}
\end{figure}

\subsection{Outcome extraction}

We apply regex-based pattern matching to extract the outcome of each case (rejection upheld, rejection reversed, or remanded). For the few cases (three across RAB-Cred) whose outcome was not automatically determined via pattern matching, we labeled the outcome manually.

\begin{figure}[h]
\begin{lstlisting}[language=python]
MONTHS_DA = r'(januar|februar|marts|april|maj|juni|juli|august|september|oktober|november|december)'
YEAR = r'\d{4}'
IMMIGRATION_SERVICE = r'Udlænding(?:estyrelsens|eservice|eservices|estyrelsen)'

rejection_upheld_patterns = [
    rf'Flertallet stemte derfor for at tiltræde Udlændingestyrelsens afgørelse',
    rf'Nævnet stadfæstede i {MONTHS_DA} {YEAR} {IMMIGRATION_SERVICE} afgørelse',
    rf'Flygtningenævnet stadfæster derfor {IMMIGRATION_SERVICE} afgørelse',
    rf'stadfæster Flygtningenævnet derfor {IMMIGRATION_SERVICE} afgørels',
    rf'Nævnet stadfæstede i {MONTHS_DA}.*?{YEAR} {IMMIGRATION_SERVICE} afgørelse',
    rf'ikke betingelserne for opholdstilladelse',
    rf'ikke sandsynliggjort, at ansøgeren.*?vil risikere forfølgelse',
    rf'fandt.*?ikke, at ansøgeren havde krav på opholdstilladelse',
    rf'ikke, at ansøgerens.*?ville være i en sådan risiko herfor, at der var grundlag for at meddele asyl',
    rf'ikke,.*?at ansøgerne skulle meddeles opholdstilladelse',
    rf'kan det ikke antages, at ansøgeren ved en tilbagevenden skulle være i en reel risiko',
    rf'Flygtningenævnets afslag',
    rf'ikke.*?at den kan begrunde opholdstilladelse efter udlændingelovens',
    rf'Flygtningenævnet finder.*?ikke, at ansøgeren.*?risikerer',
    rf'ikke, at ansøgeren.*?ville være i risiko',
    rf'ikke fandtes at være asylbegrundende',
    rf'Flygtningenævnet fandt.*?ikke, at ansøgeren ved en tilbagevenden.*?ville risikere',
    rf'ikke, at ansøgeren opfylder betingelserne for asyl',
    rf'opfyldte ansøgeren ikke betingelserne for at få asyl',
    rf'han ikke kunne påberåbe sig den beskyttelse, som følger af udlændingelovens',
    rf'Flygtningenævnet fandt ikke, at disse forhold kunne begrunde',
    rf'meddeler derfor ansøgeren afslag på opholdstilladelse',
    rf'stadfæster Flygtningenævnet {IMMIGRATION_SERVICE} afgørelse',
    rf'stadfæster herefter {IMMIGRATION_SERVICE} afgørelse',
    rf'stadfæstede i (?:{MONTHS_DA}) {YEAR} {IMMIGRATION_SERVICE} afgørelse',
    rf'finder Flygtningenævnet heller ikke,? at det vil være uproportionalt.*?opholdstilladelse'
]
rejection_reversed_patterns = [
    rf'Klageren opfylder således betingelserne for at blive meddelt opholdstilladelse',
    rf'Nævnet meddelte i opholdstilladelse(.*?)til',
    rf'Nævnet meddelte i (?:{MONTHS_DA}) {YEAR} opholdstilladelse',
    rf'Flygtningenævnet ophævede derfor {IMMIGRATION_SERVICE}',
    rf'Flygtningenævnet ændrer derfor {IMMIGRATION_SERVICE} afgørelse',
    rf'finder Flygtningenævnet således, at De skal meddeles opholdstilladelse',
    rf'Nævnet omgjorde i (?:{MONTHS_DA}) {YEAR} {IMMIGRATION_SERVICE} afgørelse',
    rf'Nævnet genoptog og omgjorde i (?:{MONTHS_DA}) {YEAR}',
    rf'besluttet at genoptage sagen og omgøre {IMMIGRATION_SERVICE} afgørelse',
    rf'meddeler derfor klageren opholdstilladelse',
    rf'Nævnet ændrede i (?:{MONTHS_DA}) {YEAR} {IMMIGRATION_SERVICE} afgørelse',
    rf'klageren meddeles opholdstilladelse efter udlændingelovens'
]
remanded_patterns = [
    rf'Nævnet hjemviste i',
    rf'sagen bør hjemvises til {IMMIGRATION_SERVICE}',
    rf'Nævnet hjemviste i (?:{MONTHS_DA}) {YEAR} {IMMIGRATION_SERVICE} afgørelse'
]
\end{lstlisting}
\caption{Regex patterns used to label the outcome of each case.}
\end{figure}

\section{Experimental set-up}

Here we present the prompt variants and models used in our experiments, along with implementation details for text classification.

\subsection{Prompt variants}~\label{app:prompt-variants}

\subsubsection{System prompts}

System prompts follow a nested structure, where SP2, SP3, SP4 and SP5 extend SP1. Furthermore, SP4 extends SP3 with edge cases.

\begin{tcolorbox}[width=\linewidth, sharp corners=all, colback=white!95!black,title=(SP0) No system prompt,label=box:sp0]

\end{tcolorbox}

\begin{tcolorbox}[width=\linewidth, sharp corners=all, colback=white!95!black,title=(SP1) Persona, label=box:sp1]

 \normalsize{You are a highly experienced legal scholar and evidence analyst specializing in qualitative coding of Danish asylum cases. You have in-depth knowledge and understanding of asylum decision-making in Denmark and the crucial role of risk and credibility. You are thorough, analytical and able to dissect complex legal reasonings. 

    \vspace{\promptspacing}
    
    You will analyze asylum decision summaries to identify the presence and the sentiment of a credibility assessment. }

\end{tcolorbox}

\phantomsection
\begin{tcolorbox}[width=\linewidth, sharp corners=all, colback=white!95!black,title=(SP2) Persona + Codebook, label=box:sp2]

\normalsize{\textbf{(SP1)}

    \vspace{\promptspacing}
    
    \#\# Context: What is a credibility assessment? 

    \vspace{\promptspacing}
    
    The purpose of credibility assessment is to determine if the claimants account can be accepted as true (in whole or in part) or not. The focus is on the past and present facts asserted by the claimant. The credibility assessment relates to establishing facts, not whether these facts justify protection or whether there is a risk of persecution.  

    \vspace{\promptspacing}
    
    Notably, it is different from the risk assessment, which is conducted to determine if there, based on the established facts, is a risk of persecution or ill treatment, reaching the legal threshold for protection. This assessment is forward looking considering what would happen if the claimant was returned.  

    \vspace{\promptspacing}
    
    Code no credibility assessment in the absence of a credibility assessment. 

    \vspace{\promptspacing}
    
    \#\# Context: Credibility sentiment:  

    \vspace{\promptspacing}
    
    The overall credibility sentiment captures the decision-maker’s overall conclusion on whether the information provided by the claimant can be accepted as facts or not. It reflects the net outcome, not the presence of isolated inconsistencies or doubts. 

    \vspace{\promptspacing}
    
    Code credibility as overall positive where a credibility assessment is present and the decision-maker accepts the core of the claimant’s account as true, even if some details are questioned, discounted, or left unresolved. The risk assessment can proceed based on the information provided by the claimant, also when there are minor doubts. The acceptance of facts may be argued or it may simply be stated that the Board accepts the motive and then proceed to a risk assessment. Please distinguish between when the board accepts the facts and proceed to risk assessment and when the board notes that even if the board accepted the facts, the risk assessment would be negative.  

    \vspace{\promptspacing}
    
    Code credibility as overall negative where a credibility assessment is present and where the decision-maker rejects the core of the claimant’s account, such that the presented factual basis of the claim is not accepted as factual. Some facts may be accepted, but the central aspects motivating protection are rejected. The risk assessment proceeds without taking the central elements of the information provided by the claimant into the account, even if parts of claim are accepted. }

\end{tcolorbox}

\begin{tcolorbox}[width=\linewidth, sharp corners=all, colback=white!95!black,title=(SP3) Persona + Computer scientist prompt, label=box:sp3]
    \normalsize{\textbf{(SP1)}

\vspace{\promptspacing}

\#\# Context: What is a credibility assessment? 

\vspace{\promptspacing}

Credibility Assessment: Evaluation of whether the claimant's factual account can be accepted. This establishes facts about past/present events. This is BACKWARD-LOOKING (what happened).  

\vspace{\promptspacing}

NOT Credibility Assessment: 

- Risk assessment: Forward-looking evaluation (will persecution occur if returned?) 

- Legal threshold analysis: Whether proven facts meet asylum criteria 

- Explicitly bypassing credibility assessment: Even if the claimant’s explanation were accepted, it would not qualify for asylum. 

\vspace{\promptspacing}
 
\#\# Context: Classification categories / codes 

\vspace{\promptspacing}

"NO CREDIBILITY ASSESSMENT" - Credibility not assessed 

- The decision proceeds directly to risk/legal analysis 

- Or there are references to credibility without establishing the credibility sentiment, as the credibility of presented facts is not considered relevant for the asylum decision 

- Danish phrases indicating this: "uanset om... lægges til grund " 

\vspace{\promptspacing}

"POSITIVE CREDIBILITY ASSESSMENT" - Credibility assessed, core account ACCEPTED as true 

- Risk assessment proceeds WITH the claimant's information 

- Minor doubts about peripheral details don't change this 

- Core of the claimant’s account accepted even if some inconsistencies noted 

- Danish phrases: "forklaring(en) ... lægges til grund", "ansøgers forklaring er troværdig" 

\vspace{\promptspacing}

"NEGATIVE CREDIBILITY ASSESSMENT" - Credibility assessed, core account REJECTED 

- Risk assessment proceeds WITHOUT central elements of claimant's account 

- Core of the claimant’s account not believed, even if some peripheral facts accepted 

- Danish phrases: "forklaring(en) ... ikke lægges til grund", "ikke troværdig", "divergerende", “konstrueret”, “usandsynlig” 

\vspace{\promptspacing}

IMPORTANT NOTE: Case outcome is different from Credibility assessment 

- Asylum claim can be REJECTED despite facts being believed (positive credibility assessment)

- Asylum claim can meet the legal threshold for asylum and thus be ACCEPTED despite the claimant’s account being deemed non-credible (negative credibility assessment). }
\end{tcolorbox}

\phantomsection

\begin{tcolorbox}[width=\linewidth, sharp corners=all, colback=white!95!black,title=(SP4) Persona + Computer scientist prompt, label=box:sp4]
     \normalsize{\textbf{(SP3)}
 
\#\# Context: critical edge cases

1. Hypothetical Construction = "NO CREDIBILITY ASSESSMENT" 

   - "Even if X was accepted, it wouldn't meet the legal threshold of protection..." 
   
   - This is NOT evaluating if X is true, just analyzing consequences IF it were true 

\vspace{\promptspacing}

2. Minor Doubts + Core Accepted as True = "POSITIVE CREDIBILITY ASSESSMENT" 

   - "Some inconsistencies about dates, but the political involvement is accepted..." 
   
   - The core claim is believed despite peripheral doubts 

\vspace{\promptspacing}

3. Some Facts Accepted + Core Not Believed = "NEGATIVE CREDIBILITY ASSESSMENT" 

   - "X is accepted, but not the claimed persecution..." 
   
   - Peripheral facts accepted but asylum grounds rejected 
 }
\end{tcolorbox}

\begin{tcolorbox}[width=\linewidth, sharp corners=all, colback=white!95!black,title=(SP5) Persona + Domain expert prompt, label=box:sp5]
     \normalsize{\textbf{(SP1)}
     
\vspace{\promptspacing}
BACKGROUND: Credibility assessment (troværdighedsvurdering) refers to evaluations of whether the asylum seeker's account is believable.  

\vspace{\promptspacing}

\#\# Context: CRITICAL DISTINCTIONS: 

NEUTRAL REPORTING vs. CREDIBILITY ASSESSMENT e.g.,: 

- NOT credibility: ’Ansøgeren har henvist til...’ (The applicant has referred to...) 

- NOT credibility: ’Ansøgeren har forklaret...’ (The applicant has explained...) 

- IS credibility: ’Flygtningenævnet lægger i det væsentlige ansøgerens forklaring til grund’ OR ’Flygtningenævnet lægger ikke ansøgerens forklaring til grund’ 

- IS credibility: ’Nævnet finder ansøgerens forklaring troværdig’ (evaluation) 

- IS credibility: ’Ansøgerens forklaring er ikke sammenhængende’ (consistency assessment) 

- IS credibility: ’Ansøægers forklaring fremstår selv-oplevet og detaljeret’ OR ’ ansøgers forklaring fremstår ikke selv-oplevet og detaljeret’ (detail assessment and genuineness assessment) 

\vspace{\promptspacing}

\#\# Context: LEGAL ASSESSMENT vs. CREDIBILITY ASSESSMENT: 

THE KEY DISTINCTION: Legal assessments evaluate whether PROVEN FACTS meet legal criteria. Credibility assessments evaluate whether the applicant’s TESTIMONY is believable. 

- NOT credibility: Legal outcome formulations, evaluating whether events meet legal thresholds for asylum 

- NOT credibility: Assessments of severity, intensity, or legal thresholds (e.g., ’disse forhold har haft et sådan omfang og intensitet’) 

- NOT credibility: Risk assessments based on country conditions or objective circumstances 

- IS credibility: Evaluating whether the applicant’s ACCOUNT of events is believable 

- IS credibility: Assessing the truthfulness or consistency of the applicant’s TESTIMONY 

\vspace{\promptspacing}

\#\# Context: CORE PRINCIPLE: When analyzing a case, ask yourself the key question: ’Is the decision-maker assessing whether something happened, and/or the applicant’s fear/convictions/political opinions/sexual orientation are genuine [IS CREDIBILITY], OR are they evaluating the risk and legal implications of accepted facts? [NOT CREDIBILITY]’ 

 }
\end{tcolorbox}

\subsubsection{User prompts}

\begin{tcolorbox}[width=\linewidth, sharp corners=all, colback=white!95!black,title=(UP1), label=box:UP1]
    \normalsize{Analyze the following legal case decision from the Flygtningenævnet (Danish Refugee Appeals Board) and classify it into one of the following three categories: [NO CREDIBILITY ASSESSMENT, POSITIVE CREDIBILITY ASSESSMENT, NEGATIVE CREDIBILITY ASSESSMENT]. Your answer should only contain the chosen category in capital letters. 
    
\vspace{\promptspacing}

Case text: \textbf{[INSERT CASE TEXT HERE]}

\vspace{\promptspacing}

Category: }
\end{tcolorbox}

\begin{tcolorbox}[width=\linewidth, sharp corners=all, colback=white!95!black,title=(UP2),label=box:UP2]
    \normalsize{Analyze the following legal case decision from the Flygtningenævnet (Danish Refugee Appeals Board) and determine whether credibility is assessed.

\vspace{\promptspacing}

Case text: \textbf{[INSERT CASE TEXT HERE]}

\vspace{\promptspacing}

Is there a credibility assessment? Your answer should only contain "Y" or "N". 
}

\rule{\textwidth}{2pt}

\textit{If the answer is "N", stop here.}

\textit{If the answer is "Y", proceed to the follow-up prompt:}

\rule{\textwidth}{2pt}

Now, determine the sentiment of the credibility assessment. Is the credibility assessment POSITIVE or NEGATIVE? Your answer should only contain the chosen category in capital letters. 

\end{tcolorbox}

\phantomsection
\begin{tcolorbox}[width=\linewidth, sharp corners=all, colback=white!95!black,title=(UP1-FS) Few-shot,label=box:UP1-FS]
    \normalsize{Analyze the following legal case decision from the Flygtningenævnet (Danish Refugee Appeals Board) and classify it into one of the following three categories: [NO CREDIBILITY ASSESSMENT, POSITIVE CREDIBILITY ASSESSMENT, NEGATIVE CREDIBILITY ASSESSMENT]. Your answer should only contain the chosen category in capital letters. 

\vspace{\promptspacing}

Case text: \textbf{[INSERT EXAMPLE 1 HERE]}

\vspace{\promptspacing}

Category: NO CREDIBILITY ASSESSMENT

\vspace{\promptspacing}

Case text:\textbf{ [INSERT EXAMPLE 2 HERE]}

\vspace{\promptspacing}

Category: POSITIVE CREDIBILITY ASSESSMENT

\vspace{\promptspacing}

Case text: \textbf{[INSERT EXAMPLE 3 HERE]}

\vspace{\promptspacing}

Category: NEGATIVE CREDIBILITY ASSESSMENT

\vspace{\promptspacing}

Case text: \textbf{[INSERT CASE TEXT HERE]}

\vspace{\promptspacing}

Category: }
\end{tcolorbox}

\begin{tcolorbox}[width=\linewidth, sharp corners=all, colback=white!95!black,title=(UP3) Zero-shot Chain-of-Thought,label=box:UP3]
    \normalsize{Analyze the following legal case decision from the Flygtningenævnet (Danish Refugee Appeals Board) and classify it into one of the following three categories: [NO CREDIBILITY ASSESSMENT, POSITIVE CREDIBILITY ASSESSMENT, NEGATIVE CREDIBILITY ASSESSMENT].  

\vspace{\promptspacing}

Case text: \textbf{[INSERT CASE TEXT HERE] }

\vspace{\promptspacing}

Let’s think step-by-step.}

\rule{\textwidth}{2pt}

To conclude, classify the case into one of the following three categories: [NO CREDIBILITY ASSESSMENT, POSITIVE CREDIBILITY ASSESSMENT, NEGATIVE CREDIBILITY ASSESSMENT]. Your answer should only contain the chosen category in capital letters. 

\end{tcolorbox}

\begin{tcolorbox}[width=\linewidth, sharp corners=all, colback=white!95!black,title=(UP4) Zero-shot Metacognitive,label=box:UP4]
    \normalsize{Analyze the following legal case decision from the Flygtningenævnet (Danish Refugee Appeals Board) and classify it into one of the following three categories: [NO CREDIBILITY ASSESSMENT, POSITIVE CREDIBILITY ASSESSMENT, NEGATIVE CREDIBILITY ASSESSMENT].  

\vspace{\promptspacing}

Case text:\textbf{ [INSERT CASE TEXT HERE] }

\vspace{\promptspacing}

As you perform this task, follow these steps: 

1. Clarify your understanding of the task and the case. 

2. Make a preliminary judgment on whether a credibility assessment is made in the case, and if so, its sentiment. 

3. Critically assess your preliminary analysis. If you are unsure about your initial judgment, reassess it. 

4. Confirm your final answer and explain the reasoning behind your decision. 

5. Evaluate your confidence (0-100\%) in your analysis and provide an explanation for this confidence level.}

\rule{\textwidth}{2pt}

To conclude, classify the case into one of the following three categories: [NO CREDIBILITY ASSESSMENT, POSITIVE CREDIBILITY ASSESSMENT, NEGATIVE CREDIBILITY ASSESSMENT]. Your answer should only contain the chosen category in capital letters.  

\end{tcolorbox}

\vspace{1em}

\paragraph{Few-shot examples}\label{app:fewshotexamples}

The following validation set cases are used as few-shot examples:
\begin{itemize}[itemsep=0pt,parsep=0pt]
    \item EXAMPLE 1: No credibility assessmwent: ID 19715 \url{https://fln.dk/praksis/2020/december/syri202051/}
    \item EXAMPLE 2: Positive credibility assessment: ID 18571 \url{https://fln.dk/praksis/2019/november/egyp20194/}
    \item EXAMPLE 3: Negative credibility assessment: ID 20194 \url{https://fln.dk/praksis/2021/februar/syri202118/}
\end{itemize}

\vspace{1em}

\begin{casebox}[title={\textbf{Example 1: No Credibility Assessment~~~~~\scriptsize (Case text translated with DeepL.com )}}]
\footnotesize In August 2020, the Board upheld the Danish Immigration Service’s decision regarding a female citizen of Syria. She arrived in 2016. The Refugee Board stated: “The complainant is an ethnic Arab from Hajar Al-Aswad, Damascus, Syria. The appellant has been politically active on Facebook, where she shares critical posts about conditions in Syria. The case file shows that the appellant entered Denmark in [the summer] of 2016 and that she was granted a residence permit in [the fall] of 2019 pursuant to Section 7(3) of the Aliens Act. In the fall of 2020, the complainant appealed to the Refugee Board against the Danish Immigration Service’s decision, claiming a right to a residence permit pursuant to Section 7(1) of the Aliens Act. In support of this, the complainant has stated that she fears the Syrian authorities will subject her to violence, kidnapping, and rape because her family is wanted. The complainant has stated that her maternal uncles have been arrested, killed, or forced to flee to Jordan because one of her maternal uncles had smashed the president’s statue, and photos of him with his foot on the statue have been circulated. After the Syrian authorities burned the complainant’s maternal uncles’ houses and motorcycles, the complainant fled with her mother and sisters to Jordan. The complainant’s maternal aunts have been arrested and are being punished by the Syrian regime. The complainant’s father and several of her father’s family members have been granted residence permits under Section 7(1) of the Aliens Act. In support of her asylum claim, the complainant has further stated that she fears the Syrian authorities because she is politically active on Facebook. The complainant has stated in this regard that, should she return, she fears the war and the bombings. The complainant has stated that she fears the government, the president, and the military. The complainant has stated that she uses Facebook to share information about the war. The complainant primarily writes the posts she shares herself. The complainant primarily shares posts with political content, as they concern the situation in Syria. The complainant’s Facebook profile is public, as she wants everyone to know what is happening. The complainant began sharing posts about Syria in late 2018. In the event of a return to Syria, the complainant fears that the authorities will search her phone and arrest her based on her Facebook activism. The Refugee Board finds—like the Danish Immigration Service—that it is based on the complainant’s own assumption that her Facebook posts, which according to her are critical of the regime, have come to the attention of the Syrian authorities. In this regard, the Refugee Board has emphasized that the complainant has a limited number of “friends” on Facebook, and that these consist solely of friends and family. Furthermore, and since the complainant—contrary to what was stated in a concurrent decision in her mother’s case— be at risk due to her kinship with her uncle, the Refugee Board does not find that the appellant has established a likelihood that, upon return to Syria, she would be at risk of persecution or abuse covered by Section 7(1) or (2) of the Aliens Act. The appellant thus does not meet the conditions for being granted a residence permit pursuant to Section 7(1) or Section 7(2) of the Aliens Act, and therefore the Refugee Board upholds the Danish Immigration Service’s decision.” Syri/2020/51/EDO

\hrulefill

\vspace{2mm}

\normalsize \textbf{Annotators H1 and H2}: Q1: Credibility assessment present? \textbf{No} (Confidence: High)

\vspace{2mm}
\textbf{Outcome}: Rejection upheld

\end{casebox}

\vspace{1em}

\begin{casebox}[title={\textbf{Example 2: Positive  Credibility Assessment~~~~~\scriptsize (Case text translated with DeepL.com )}}]
\footnotesize In October 2019, the Board upheld the Danish Immigration Service’s decision regarding a male citizen of Egypt. He entered the country in 2016. The Refugee Board stated: “The applicant is an ethnic Arab and a Coptic Christian from [town name], Dakahliya, Egypt. The applicant has not been a member of political or religious associations or organizations, nor has he otherwise been politically active. As grounds for asylum, the applicant has cited his fear of being killed by his own family or random Egyptians because he is homosexual. In support of this, he has stated that it is no longer a secret that he is homosexual, as people living here have become aware of this. The applicant’s father has learned of the applicant’s sexuality through these people and has threatened over the phone that he will cut the applicant into small pieces and have nothing to do with him. He has also cited as a basis for his asylum claim that he fears being killed by two men named [A] and [B] or by random individuals because he is Christian. In support of this, he has stated that at one point he picked up [A] and [B] in a tuk-tuk and drove them to a deserted area. [B] pulled out a knife, and the applicant assumed he intended to kill him, so he fled the scene. The applicant reported the incident to the police, who arrested [A] and [B]. During the trial in 2005, the applicant was offered money by [A]’s family to withdraw the complaint. He received an indirect threat from [A]’s maternal grandfather. The applicant subsequently dropped the case against [A] and [B], as he did not believe he would gain anything from their imprisonment. The applicant has had no problems with them since 2005. However, the applicant suspects that there is a connection between the case against [A] and [B] and the fact that the applicant’s father was wrongfully accused of various criminal offenses from 2005 to 2010. From 2006 until his departure on [date in the spring] 2016, the applicant lived in both Cairo and Sharm el-Sheikh without experiencing any problems with [A] or [B]. He has further noted that random attacks on Coptic Christians occur in Egypt. Finally, as a basis for his asylum claim, the applicant has stated that he fears the Egyptian intelligence service because he has applied for asylum in Denmark. The Refugee Board, like the Danish Immigration Service, finds the applicant’s explanation credible. However, the Refugee Board finds, for the same reasons stated by the Danish Immigration Service, that neither the general conditions for Coptic Christians in Egypt, the applicant’s fear of the Egyptian intelligence service, nor the applicant’s conflict with [A], [B], and their families can justify a residence permit under Section 7 of the Aliens Act. It appears from the available background information that it is not illegal to be an LGBT person or to have sex with a person of the same sex in Egypt. Notwithstanding that the available background information also indicates that in September 2017 the authorities adopted a tougher stance toward homosexuals and that, in connection with this, a bill was introduced to criminalize homosexuality, the Refugee Board does not find that the general conditions —despite being difficult—for homosexuals in Egypt can in themselves lead to the applicant being granted a residence permit. Nor does the majority of the Refugee Board find that the applicant has substantiated that he has been specifically and individually persecuted as a result of his sexuality. In this regard, the majority has emphasized that, based on the information regarding the telephone threat from the applicant’s father, it has not been substantiated that the father will actively seek out the applicant and carry out his threat. The majority thus finds that the applicant can be directed to reside outside his father’s home region, including in Cairo, where he has been able to reside without issue for approximately 4 years starting in 2005. The fact that, according to the applicant’s account, other individuals in Egypt have become aware of his sexuality cannot lead to a different assessment. The majority of the Refugee Board thus finds, after a comprehensive assessment, that the applicant has not established that he would be at risk of persecution or abuse covered by Section 7 of the Aliens Act upon a return to Egypt. The Refugee Board therefore upholds the decision of the Danish Immigration Service.” Egyp/2019/4/DH

\hrulefill

\vspace{2mm}

\normalsize \textbf{Annotators H1 and H2}: Q1: Credibility assessment present? \textbf{Yes} (Confidence: High)

\textbf{Annotators H1 and H2}: Q2: Credibility assessment sentiment \textbf{Positive} (Confidence: High)

\vspace{2mm}
\textbf{Outcome}: Rejection upheld

\end{casebox}

\vspace{1em}

\begin{casebox}[title={\textbf{Example 3: Negative Credibility Assessment~~~~~\scriptsize (Case text translated with DeepL.com )}}]
\footnotesize{In February 2021, the Board overturned the Danish Immigration Service’s decision regarding a female citizen of Syria. She entered the country in 2016. The Refugee Board stated: “The appellant is an ethnic Kurd and a Sunni Muslim from Dirik, Al Hasakah, Syria. The complainant has not been a member of any political or religious associations or organizations, nor has she been politically active in any other way. The Danish Immigration Service granted the appellant a residence permit in the summer of 2016 pursuant to Section 7(3) of the Aliens Act. During the original asylum case, the appellant cited as her reason for seeking asylum that she feared the general conditions in Syria if she were to return. In support of this, the complainant had stated that there was a war in Syria and that the regular bombings in Syria were affecting her mental state. The complainant had further stated that she wished to live a peaceful life with her children. In the summer of 2020, the Danish Immigration Service decided to refuse to extend the complainant’s residence permit pursuant to Section 11(2), second sentence, of the Aliens Act. The Danish Immigration Service has assessed that the basis for the complainant’s residence permit no longer exists. The complainant has continued to cite general conditions as grounds for asylum, as well as her family’s political activities and conflicts with the authorities. For the reasons stated by the Danish Immigration Service, the Refugee Board agrees that the complainant has not demonstrated that, upon return to Syria, she would be at risk of persecution or abuse covered by Section 7(1) or (2) of the Aliens Act. The complainant’s statements to the Refugee Board do not warrant a different assessment. In this regard, the Refugee Board has emphasized that the complainant’s statements cannot, in essence, be relied upon, as the complainant has, among other things, provided detailed accounts of her spouse’s, sons’, and sons-in-law’s political activities, and that, according to a neighbor, the family’s home was visited by the authorities with the intention of arresting the family shortly before their departure. The fact that the complainant and family members in Denmark participated in a demonstration cannot lead to a different assessment either, as the complainant cannot be presumed to have come under the scrutiny of the Syrian authorities for that reason, regardless of whether photos of the demonstration may have been posted on Facebook. The Refugee Board further agrees with the Danish Immigration Service that the general conditions in Damascus are no longer such that anyone would be at a real risk of being subjected to abuse in violation of Article 3 of the European Convention on Human Rights solely as a result of their mere presence in the area. As stated by the Danish Immigration Service, the Refugee Board must now determine whether refusing to extend the complainant’s residence permit would violate Article 8 of the European Convention on Human Rights. The complainant is Kurdish and 63 years old. She left Syria around 2012 and entered Denmark in 2016. The complainant is a single woman, illiterate, and has never been part of the labor market. The complainant has nine adult children, one of whom lives in Sweden, while the rest live in Denmark. The complainant has never lived alone and, since her arrival in Denmark, has lived with one of her adult sons and his family; the complainant’s youngest son, aged 22, now also lives with her. Furthermore, based on the complainant’s statement, the Refugee Board finds that she has absolutely no family or other network in Syria. On that basis and following a comprehensive and concrete assessment, the Refugee Board finds that it would currently be contrary to Denmark’s international obligations, cf. Article 8 of the European Convention on Human Rights, to refuse to extend the complainant’s residence permit. The Refugee Board therefore extends the complainant’s residence permit pursuant to Section 7(3) of the Aliens Act.” Syri/2021/18/HZC}

\hrulefill

\vspace{2mm}

\normalsize \textbf{Annotators H1 and H2}: Q1: Credibility assessment present? \textbf{Yes} (Confidence: High)

\textbf{Annotators H1 and H2}: Q2: Credibility assessment sentiment \textbf{Negative} (Confidence: High)

\vspace{2mm}
\textbf{Outcome}: Rejection reversed

\end{casebox}

%\captionof{table}{ Translation of three case examples used in the few-shot prompting experiment DeepL.com (free version).}

\clearpage
\subsection{Model selection and implementation}\label{app:model_selection}

\subsubsection{Model selection}

We evaluate the following 21 models on the validation set. The top-5 models which are selected for evaluation on the test set are in bold. All models are pulled from HuggingFace and used in their default precision.

\begin{table}[h]
\centering
\resizebox{\textwidth}{!}{%
\begin{tabular}{m{1.3cm}m{8cm}m{1cm}m{13cm}}
\toprule
 \multicolumn{2}{c}{\textbf{huggingface model}} & \textbf{context length} & \textbf{multilingual capabilities mentioned in Hugging Face model card} \\ \midrule

\multirow{2}{*}{Aya}  & & & \multirow{5}{=}{"Languages covered: The model is particularly optimized for multilinguality and supports the following languages: Arabic, Chinese (simplified \& traditional), Czech, Dutch, English, French, German, Greek, Hebrew, Hindi, Indonesian, Italian, Japanese, Korean, Persian, Polish, Portuguese, Romanian, Russian, Spanish, Turkish, Ukrainian, and Vietnamese"} \\
 & \href{https://huggingface.co/CohereLabs/aya-expanse-8b}{CohereLabs/aya-expanse-8b} & 8K &  \\
 & & & \\
 & \href{https://huggingface.co/CohereLabs/aya-expanse-32b}{CohereLabs/aya-expanse-32b} & 128K &  \\
 & & & \\
 
\midrule
\multirow{3}{*}{Gemma} & \href{https://huggingface.co/google/gemma-3-27b-it}{\textbf{gemma-3-27b-it}} & \multirow{3}{*}{128K} & \multirow{3}{=}{"multilingual support in over 140 languages" "The training dataset includes content in over 140 languages."} \\
 & \href{https://huggingface.co/google/gemma-3-12b-it}{gemma-3-12b-it} &  &  \\
 & \href{https://huggingface.co/google/gemma-3-4b-it}{gemma-3-4b-it} &  &  \\ \midrule
Granite & \href{https://huggingface.co/ibm-granite/granite-4.0-micro}{ibm-granite/granite-4.0-micro} & 128K & "Supported Languages: English, German, Spanish, French, Japanese, Portuguese, Arabic, Czech, Italian, Korean, Dutch, and Chinese. Users may finetune Granite 4.0 models for languages beyond these languages." \\ \midrule

\multirow{2}{*}{Llama} & \href{https://huggingface.co/meta-llama/Llama-3.1-8B-Instruct}{meta-llama/Llama-3.1-8B-Instruct} & 128K & "Supported languages: English, German, French, Italian, Portuguese, Hindi, Spanish, and Thai." "Note: Llama 3.1 has been trained on a broader collection of languages than the 8 supported languages. Developers may fine-tune Llama 3.1 models for languages beyond the 8 supported languages " \\ \cmidrule{2-4} 

 & \href{https://huggingface.co/meta-llama/Llama-3.2-3B-Instruct}{meta-llama/Llama-3.2-3B-Instruct} & 128K & "Supported Languages: English, German, French, Italian, Portuguese, Hindi, Spanish, and Thai are officially supported. Llama 3.2 has been trained on a broader collection of languages than these 8 supported languages. Developers may fine-tune Llama 3.2 models for languages beyond these supported languages" \\ \midrule
 
\multirow{2}{*}{Phi} & \href{https://huggingface.co/microsoft/phi-4}{\textbf{microsoft/phi-4}} & 16K & "Multilingual data constitutes about 8\% of our overall data. " "The model is trained primarily on English text. Languages other than English will experience worse performance." "phi-4 is not intended to support multilingual use. " \\ \cmidrule{2-4} 
 & \href{https://huggingface.co/microsoft/Phi-4-mini-instruct}{microsoft/Phi-4-mini-instruct} & 128K & "Supported languages: Arabic, Chinese, Czech, Danish, Dutch, English, Finnish, French, German, Hebrew, Hungarian, Italian, Japanese, Korean, Norwegian, Polish, Portuguese, Russian, Spanish, Swedish, Thai, Turkish, Ukrainian" "The model is intended for broad multilingual commercial and research use." "The Phi models are trained primarily on English text and some additional multilingual text. Languages other than English will experience worse performance as well as performance disparities across non-English." \\ \midrule
 
\multirow{3}{*}{Mistral} & \href{https://huggingface.co/mistralai/Ministral-3-8B-Instruct-2512}{mistralai/Ministral-3-8B-Instruct-2512} & \multirow{2}{*}{256K} & \multirow{2}{=}{"Supports dozens of languages, including English, French, Spanish, German, Italian, Portuguese, Dutch, Chinese, Japanese, Korean, Arabic."} \\
 & \href{https://huggingface.co/mistralai/Ministral-3-14B-Instruct-2512}{\textbf{mistralai/Ministral-3-14B-Instruct-2512}} &  &  \\ \cmidrule{2-4} 
 & \href{https://huggingface.co/mistralai/Mistral-Small-3.2-24B-Instruct-2506}{\textbf{mistralai/Mistral-Small-3.2-24B-Instruct-2506}} & 128K & "Supports dozens of languages, including English, French, German, Greek, Hindi, Indonesian, Italian, Japanese, Korean, Malay, Nepali, Polish, Portuguese, Romanian, Russian, Serbian, Spanish, Swedish, Turkish, Ukrainian, Vietnamese, Arabic, Bengali, Chinese, Farsi." \\ \midrule
 
\multirow{6}{*}{Qwen} & \href{https://huggingface.co/Qwen/Qwen2.5-32B-Instruct}{Qwen/Qwen2.5-32B-Instruct} & 256K & \multirow{4}{=}{"Multilingual support for over 29 languages, including Chinese, English, French, Spanish, Portuguese, German, Italian, Russian, Japanese, Korean, Vietnamese, Thai, Arabic, and more."} \\
 & \href{https://huggingface.co/Qwen/Qwen2.5-7B-Instruct}{Qwen/Qwen2.5-7B-Instruct} & 131K &  \\
 & \href{https://huggingface.co/Qwen/Qwen2.5-3B-Instruct}{Qwen/Qwen2.5-3B-Instruct} & 128K &  \\
 & \href{https://huggingface.co/Qwen/Qwen2.5-14B-Instruct}{Qwen/Qwen2.5-14B-Instruct} & 128K &  \\ \cmidrule{2-4} 
 & \href{https://huggingface.co/Qwen/Qwen3-30B-A3B-Instruct-2507}{\textbf{Qwen/Qwen3-30B-A3B-Instruct-2507}} & \multirow{2}{*}{256K} & \multirow{2}{*}{"Substantial gains in long-tail knowledge coverage across multiple languages."} \\
 & \href{https://huggingface.co/Qwen/Qwen3-4B-Instruct-2507}{Qwen/Qwen3-4B-Instruct-2507} &  &  \\ \midrule
Bielik & \href{https://huggingface.co/speakleash/Bielik-11B-v3.0-Instruct}{speakleash/Bielik-11B-v3.0-Instruct} & 32K & "Developed and trained on multilingual text corpora across 32 European languages, with emphasis on Polish" \\ \midrule
EuroLLM & \href{https://huggingface.co/utter-project/EuroLLM-22B-Instruct-2512}{utter-project/EuroLLM-22B-Instruct-2512} & 32K & "Language(s) (NLP): Bulgarian, Croatian, Czech, Danish, Dutch, English, Estonian, Finnish, French, German, Greek, Hungarian, Irish, Italian, Latvian, Lithuanian, Maltese, Polish, Portuguese, Romanian, Slovak, Slovenian, Spanish, Swedish, Arabic, Catalan, Chinese, Galician, Hindi, Japanese, Korean, Norwegian, Russian, Turkish, and Ukrainian. " "The EuroLLM project has the goal of creating a suite of LLMs capable of understanding and generating text in all European Union languages as well as some additional relevant languages." \\ \bottomrule
\end{tabular}%
}
\caption{Model selection, including direct links to Hugging Face model cards, and direct quotes from each model's model card related to multilingual capabilities. The 5 models which were selected for evaluation on the test set are highlighted in bold.}
\end{table}

\clearpage

\subsubsection{Decoding and constrained generation}\label{app:constrained-gen}

Following existing work~\cite{bavaresco-etal-2025-llms,tornberg2024best,pavlovic-poesio-2024-effectiveness}, we apply greedy decoding across all model-prompt combinations.  For reasoning steps (in UP3 and UP4), we initially considered using each model's default and/or explicitly recommended sampling parameters, but did not observe consistent performance improvement.

To ensure that the LLM produces a valid category in response to classification queries, we use the \texttt{outlines} library~\cite{willard2023efficientguidedgenerationlarge} wrapping around \texttt{transformers} generation. For UP1, UP1-FS and the second turn in UP3 \& UP4, we apply the following output schema:
\begin{lstlisting}[language=Python]
output_schema = Literal[
    "NO CREDIBILITY ASSESSMENT",
    "POSITIVE CREDIBILITY ASSESSMENT",
    "NEGATIVE CREDIBILITY ASSESSMENT"
]
\end{lstlisting}
For UP2, we apply the following output schema at each turn:
\begin{lstlisting}[language=Python]
# 1st turn:
output_schema = Literal["Y","N"] 

# 2nd turn
output_schema = Literal["POSITIVE","NEGATIVE"] 
\end{lstlisting}

Five models in our selection were found to have limited support for constrained generation: Mistral, EuroLLM and Bielik models. We therefore apply rule-based logic to extract the chosen category from their outputs (cf. Figure~\ref{code:classification-rule}).

\begin{figure}[h]
\begin{lstlisting}[language=Python]
if set(get_args(output_schema)) == {"Y", "N"}:
    output = output[0]
elif set(get_args(output_schema)) == {"POSITIVE", "NEGATIVE"}:
    if output[0] in ["P", "N"]:
        output = "POSITIVE" if output[0] == "P" else "NEGATIVE"
    else:
        if "POSITIVE" in output and "NEGATIVE" in output:
            if output.count("POSITIVE") > output.count("NEGATIVE"): output="POSITIVE"
            else: output="NEGATIVE"
        else:
            if "POSITIVE" in output: output="POSITIVE"
            else: output="NEGATIVE"
elif set(get_args(output_schema)) == {"POSITIVE CREDIBILITY ASSESSMENT", "NEGATIVE CREDIBILITY ASSESSMENT", "NO CREDIBILITY ASSESSMENT"}:
    if not ("POSITIVE" in output or "NEGATIVE" in output or "NO " in output):
        logging.warning(f"LLM output '{output}' could not be mapped to 3-class prediction.")
        output = "NEGATIVE CREDIBILITY ASSESSMENT"
    else:
        output = "POSITIVE CREDIBILITY ASSESSMENT" if "POSITIVE" in output else "NEGATIVE CREDIBILITY ASSESSMENT" if "NEGATIVE" in output else "NO CREDIBILITY ASSESSMENT"
\end{lstlisting}
\caption{How we extract the predicted class from the LLM's response.}\label{code:classification-rule}
\end{figure}

For reasoning steps (in UP3 and UP4), we do not constrain the content or length of the output. We set \verb|max_new_tokens| to an arbitrarily large number (100,000).

\subsubsection{Compute infrastructure}

We perform inference on local hardware (GeForce RTX 3080 with 10GB of VRAM) as well as remote compute nodes (single A40 with 48GB of VRAM, and single H100 with 80GB of VRAM), depending on the model size. Inference time per sample varied widely depending on the model and prompt combination, ranging from 0.1s to 2 minutes per sample.

\clearpage

\section{Validation set evaluation}\label{app:full-perf-val}

\begin{figure}[h!]
    \centering
    \includegraphics[width=\linewidth]{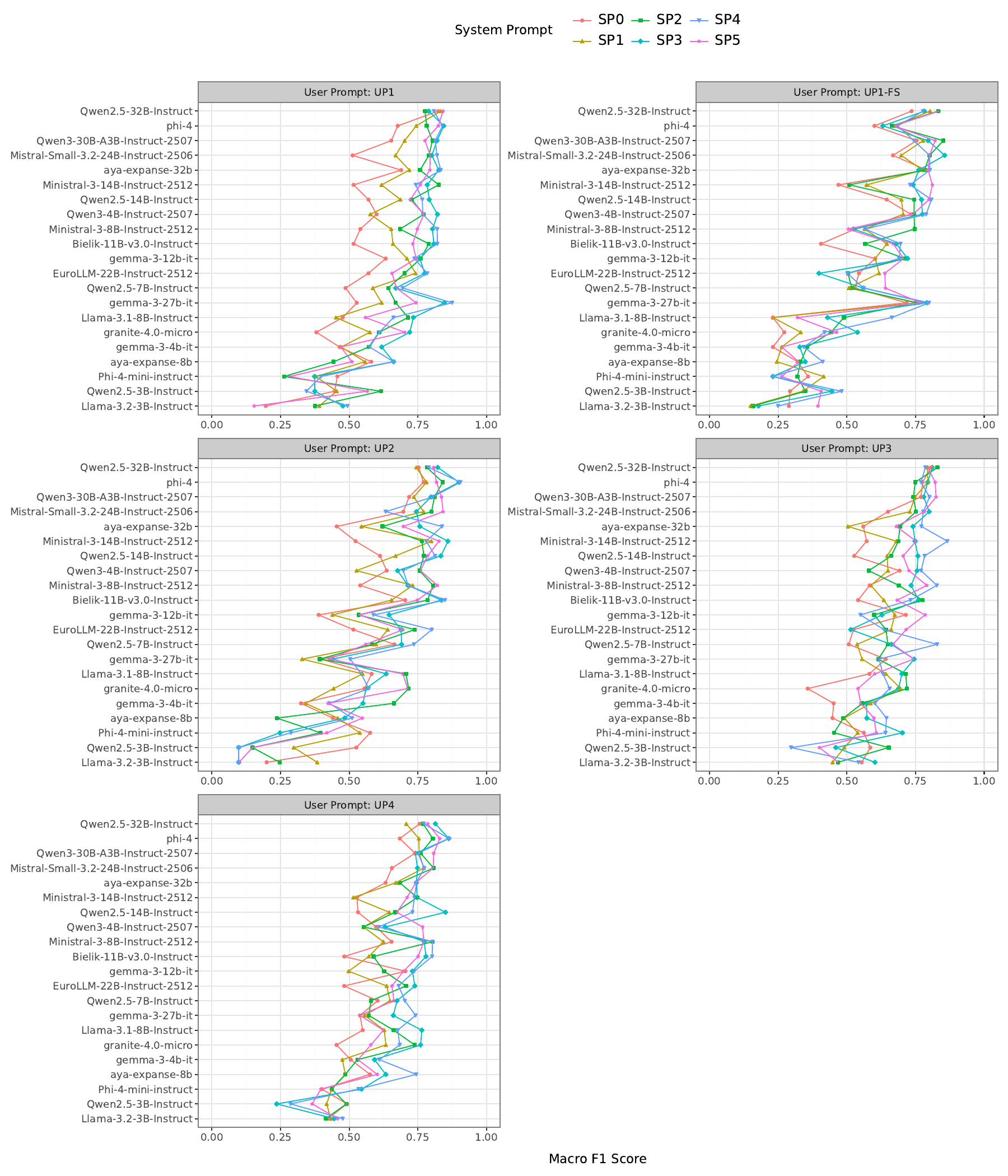}
    \caption{Classification performance of each model-prompt combination on the validation set in terms of Macro F1 (wrt. the label agreed upon by H1 and H2). Plots are split by user prompt, and color coded by system prompt.}
    \label{fig:app:f1_scores_by_model_and_prompt_variant}
\end{figure}

Figure~\ref{fig:app:f1_scores_by_model_and_prompt_variant} shows the performance of individual LLM annotators on the validation set. Each datapoint corresponds to a single model-UP-SP combination, with 630 datapoints in total.  

\clearpage

The top 15 LLM annotators are selected by averaging performance (macro F1) for the top-3 UP-SP combinations for each model, and taking the top 5 models $\times$ 3 prompts. Table~\ref{tab:app:selected-perf-val} shows the performance of the resulting selected LLM annotators on the validation set.

\begin{table*}[h]
    \centering
    \resizebox{\textwidth}{!}{%
    \begin{tabular}{cccccc}
    \toprule
Model & System Prompt & User Prompt & Macro-F1 Score (\%) & Cohen's Kappa & Accuracy (\%) \\ \midrule
phi-4 & SP4 & UP2 & {\cellcolor[HTML]{006837}} \color[HTML]{F1F1F1} 90.51 & {\cellcolor[HTML]{006837}} \color[HTML]{F1F1F1} 0.86 & {\cellcolor[HTML]{006837}} \color[HTML]{F1F1F1} 91.43 \\
phi-4 & SP3 & UP2 & {\cellcolor[HTML]{0C7F43}} \color[HTML]{F1F1F1} 89.99 & {\cellcolor[HTML]{06733D}} \color[HTML]{F1F1F1} 0.86 & {\cellcolor[HTML]{006837}} \color[HTML]{F1F1F1} 91.43 \\
phi-4 & SP4 & UP4 & {\cellcolor[HTML]{D5ED88}} \color[HTML]{000000} 86.34 & {\cellcolor[HTML]{A2D76A}} \color[HTML]{000000} 0.81 & {\cellcolor[HTML]{B7E075}} \color[HTML]{000000} 88.57 \\
\midrule
gemma-3-27b-it & SP4 & UP1 & {\cellcolor[HTML]{9BD469}} \color[HTML]{000000} 87.50 & {\cellcolor[HTML]{69BE63}} \color[HTML]{F1F1F1} 0.83 & {\cellcolor[HTML]{4BB05C}} \color[HTML]{F1F1F1} 90.00 \\
gemma-3-27b-it & SP3 & UP1 & {\cellcolor[HTML]{FEEFA3}} \color[HTML]{000000} 84.66 & {\cellcolor[HTML]{C3E67D}} \color[HTML]{000000} 0.80 & {\cellcolor[HTML]{B7E075}} \color[HTML]{000000} 88.57 \\
gemma-3-27b-it & SP4 & UP1-FS & {\cellcolor[HTML]{A50026}} \color[HTML]{F1F1F1} 79.91 & {\cellcolor[HTML]{A50026}} \color[HTML]{F1F1F1} 0.69 & {\cellcolor[HTML]{A50026}} \color[HTML]{F1F1F1} 82.86 \\
\midrule
Ministral-3-14B-Instruct-2512 & SP4 & UP3 & {\cellcolor[HTML]{C9E881}} \color[HTML]{000000} 86.59 & {\cellcolor[HTML]{F8FCB6}} \color[HTML]{000000} 0.78 & {\cellcolor[HTML]{FFFEBE}} \color[HTML]{000000} 87.14 \\
Ministral-3-14B-Instruct-2512 & SP3 & UP2 & {\cellcolor[HTML]{E5F49B}} \color[HTML]{000000} 85.95 & {\cellcolor[HTML]{E5F49B}} \color[HTML]{000000} 0.79 & {\cellcolor[HTML]{FFFEBE}} \color[HTML]{000000} 87.14 \\
Ministral-3-14B-Instruct-2512 & SP5 & UP2 & {\cellcolor[HTML]{FA9656}} \color[HTML]{000000} 82.69 & {\cellcolor[HTML]{FED683}} \color[HTML]{000000} 0.76 & {\cellcolor[HTML]{FDBF6F}} \color[HTML]{000000} 85.71 \\
\midrule
Mistral-Small-3.2-24B-Instruct-2506 & SP3 & UP1-FS & {\cellcolor[HTML]{EFF8AA}} \color[HTML]{000000} 85.63 & {\cellcolor[HTML]{E2F397}} \color[HTML]{000000} 0.79 & {\cellcolor[HTML]{FFFEBE}} \color[HTML]{000000} 87.14 \\
Mistral-Small-3.2-24B-Instruct-2506 & SP5 & UP2 & {\cellcolor[HTML]{FEE18D}} \color[HTML]{000000} 84.18 & {\cellcolor[HTML]{E5F49B}} \color[HTML]{000000} 0.79 & {\cellcolor[HTML]{FFFEBE}} \color[HTML]{000000} 87.14 \\
Mistral-Small-3.2-24B-Instruct-2506 & SP4 & UP1 & {\cellcolor[HTML]{F26841}} \color[HTML]{F1F1F1} 81.95 & {\cellcolor[HTML]{FDAD60}} \color[HTML]{000000} 0.74 & {\cellcolor[HTML]{EA5739}} \color[HTML]{F1F1F1} 84.29 \\
\midrule
Qwen3-30B-A3B-Instruct-2507 & SP2 & UP1-FS & {\cellcolor[HTML]{FFFDBC}} \color[HTML]{000000} 85.14 & {\cellcolor[HTML]{E5F49B}} \color[HTML]{000000} 0.79 & {\cellcolor[HTML]{FFFEBE}} \color[HTML]{000000} 87.14 \\
Qwen3-30B-A3B-Instruct-2507 & SP5 & UP2 & {\cellcolor[HTML]{FDC776}} \color[HTML]{000000} 83.61 & {\cellcolor[HTML]{FAFDB8}} \color[HTML]{000000} 0.78 & {\cellcolor[HTML]{FFFEBE}} \color[HTML]{000000} 87.14 \\
Qwen3-30B-A3B-Instruct-2507 & SP5 & UP3 & {\cellcolor[HTML]{F88950}} \color[HTML]{F1F1F1} 82.48 & {\cellcolor[HTML]{F8FCB6}} \color[HTML]{000000} 0.78 & {\cellcolor[HTML]{FFFEBE}} \color[HTML]{000000} 87.14 \\
\bottomrule
\end{tabular}%
}
    \caption{\textbf{Validation set} performance of the 15 model-prompt combinations that we select for evaluation on the test set. The performance metrics are with respect to human annotations.}
    \label{tab:app:selected-perf-val}
\end{table*}

\section{Test set evaluation}

\subsubsection{Top selected model-prompt combinations}~\label{app:selected-perf-test}

\begin{table*}[h]
    \centering
    \resizebox{\textwidth}{!}{ %   
\begin{tabular}{cccccc}
\toprule
Model & System Prompt & User Prompt & Macro-F1 Score (\%) & Cohen's Kappa & Accuracy (\%) \\
\midrule
phi-4 & SP4 & UP2 & {\cellcolor[HTML]{199750}} \color[HTML]{F1F1F1} 93.66 & {\cellcolor[HTML]{0D8044}} \color[HTML]{F1F1F1} 0.91 & {\cellcolor[HTML]{0F8446}} \color[HTML]{F1F1F1} 94.00 \\
phi-4 & SP3 & UP2 & {\cellcolor[HTML]{39A758}} \color[HTML]{F1F1F1} 93.20 & {\cellcolor[HTML]{51B35E}} \color[HTML]{F1F1F1} 0.89 & {\cellcolor[HTML]{5AB760}} \color[HTML]{F1F1F1} 93.00 \\
phi-4 & SP4 & UP4 & {\cellcolor[HTML]{006837}} \color[HTML]{F1F1F1} 94.69 & {\cellcolor[HTML]{006837}} \color[HTML]{F1F1F1} 0.91 & {\cellcolor[HTML]{006837}} \color[HTML]{F1F1F1} 94.50 \\
\midrule
gemma-3-27b-it & SP4 & UP1 & {\cellcolor[HTML]{F2FAAE}} \color[HTML]{000000} 89.89 & {\cellcolor[HTML]{CBE982}} \color[HTML]{000000} 0.86 & {\cellcolor[HTML]{CBE982}} \color[HTML]{000000} 91.50 \\
gemma-3-27b-it & SP3 & UP1 & {\cellcolor[HTML]{F8864F}} \color[HTML]{F1F1F1} 86.84 & {\cellcolor[HTML]{FED683}} \color[HTML]{000000} 0.83 & {\cellcolor[HTML]{FED481}} \color[HTML]{000000} 89.50 \\
gemma-3-27b-it & SP4 & UP1-FS & {\cellcolor[HTML]{A50026}} \color[HTML]{F1F1F1} 84.39 & {\cellcolor[HTML]{A50026}} \color[HTML]{F1F1F1} 0.78 & {\cellcolor[HTML]{A50026}} \color[HTML]{F1F1F1} 86.50 \\
\midrule
Ministral-3-14B-Instruct-2512 & SP4 & UP3 & {\cellcolor[HTML]{A2D76A}} \color[HTML]{000000} 91.63 & {\cellcolor[HTML]{A5D86A}} \color[HTML]{000000} 0.87 & {\cellcolor[HTML]{ABDB6D}} \color[HTML]{000000} 92.00 \\
Ministral-3-14B-Instruct-2512 & SP3 & UP2 & {\cellcolor[HTML]{D7EE8A}} \color[HTML]{000000} 90.62 & {\cellcolor[HTML]{DCF08F}} \color[HTML]{000000} 0.86 & {\cellcolor[HTML]{E6F59D}} \color[HTML]{000000} 91.00 \\
Ministral-3-14B-Instruct-2512 & SP5 & UP2 & {\cellcolor[HTML]{FFF1A8}} \color[HTML]{000000} 89.05 & {\cellcolor[HTML]{FFF3AC}} \color[HTML]{000000} 0.84 & {\cellcolor[HTML]{FEEC9F}} \color[HTML]{000000} 90.00 \\
\midrule
Mistral-Small-3.2-24B-Instruct-2506 & SP3 & UP1-FS & {\cellcolor[HTML]{6BBF64}} \color[HTML]{000000} 92.52 & {\cellcolor[HTML]{75C465}} \color[HTML]{000000} 0.88 & {\cellcolor[HTML]{84CA66}} \color[HTML]{000000} 92.50 \\
Mistral-Small-3.2-24B-Instruct-2506 & SP5 & UP2 & {\cellcolor[HTML]{CBE982}} \color[HTML]{000000} 90.85 & {\cellcolor[HTML]{C3E67D}} \color[HTML]{000000} 0.86 & {\cellcolor[HTML]{CBE982}} \color[HTML]{000000} 91.50 \\
Mistral-Small-3.2-24B-Instruct-2506 & SP4 & UP1 & {\cellcolor[HTML]{7AC665}} \color[HTML]{000000} 92.30 & {\cellcolor[HTML]{75C465}} \color[HTML]{000000} 0.88 & {\cellcolor[HTML]{84CA66}} \color[HTML]{000000} 92.50 \\
\midrule
Qwen3-30B-A3B-Instruct-2507 & SP2 & UP1-FS & {\cellcolor[HTML]{FEE695}} \color[HTML]{000000} 88.70 & {\cellcolor[HTML]{F2FAAE}} \color[HTML]{000000} 0.85 & {\cellcolor[HTML]{FEFFBE}} \color[HTML]{000000} 90.50 \\
Qwen3-30B-A3B-Instruct-2507 & SP5 & UP2 & {\cellcolor[HTML]{FDC574}} \color[HTML]{000000} 87.95 & {\cellcolor[HTML]{FDBF6F}} \color[HTML]{000000} 0.82 & {\cellcolor[HTML]{FDB567}} \color[HTML]{000000} 89.00 \\
Qwen3-30B-A3B-Instruct-2507 & SP5 & UP3 & {\cellcolor[HTML]{FEE695}} \color[HTML]{000000} 88.70 & {\cellcolor[HTML]{FEE491}} \color[HTML]{000000} 0.83 & {\cellcolor[HTML]{FED481}} \color[HTML]{000000} 89.50 \\
\bottomrule
\end{tabular}%
}
    \caption{\textbf{Test set performance} of the 15 model-prompt combinations that we select for evaluation on the test set. The performance metrics are with respect to human annotations, taking the majority vote between annotators.}
    \label{tab:app:selected-perf-test}
\end{table*}

\clearpage
\subsubsection{Inter-LLM agreement}\label{app:model-agreement-test}

\begin{figure}[h]
    \centering
    \includegraphics[width=0.95\linewidth]{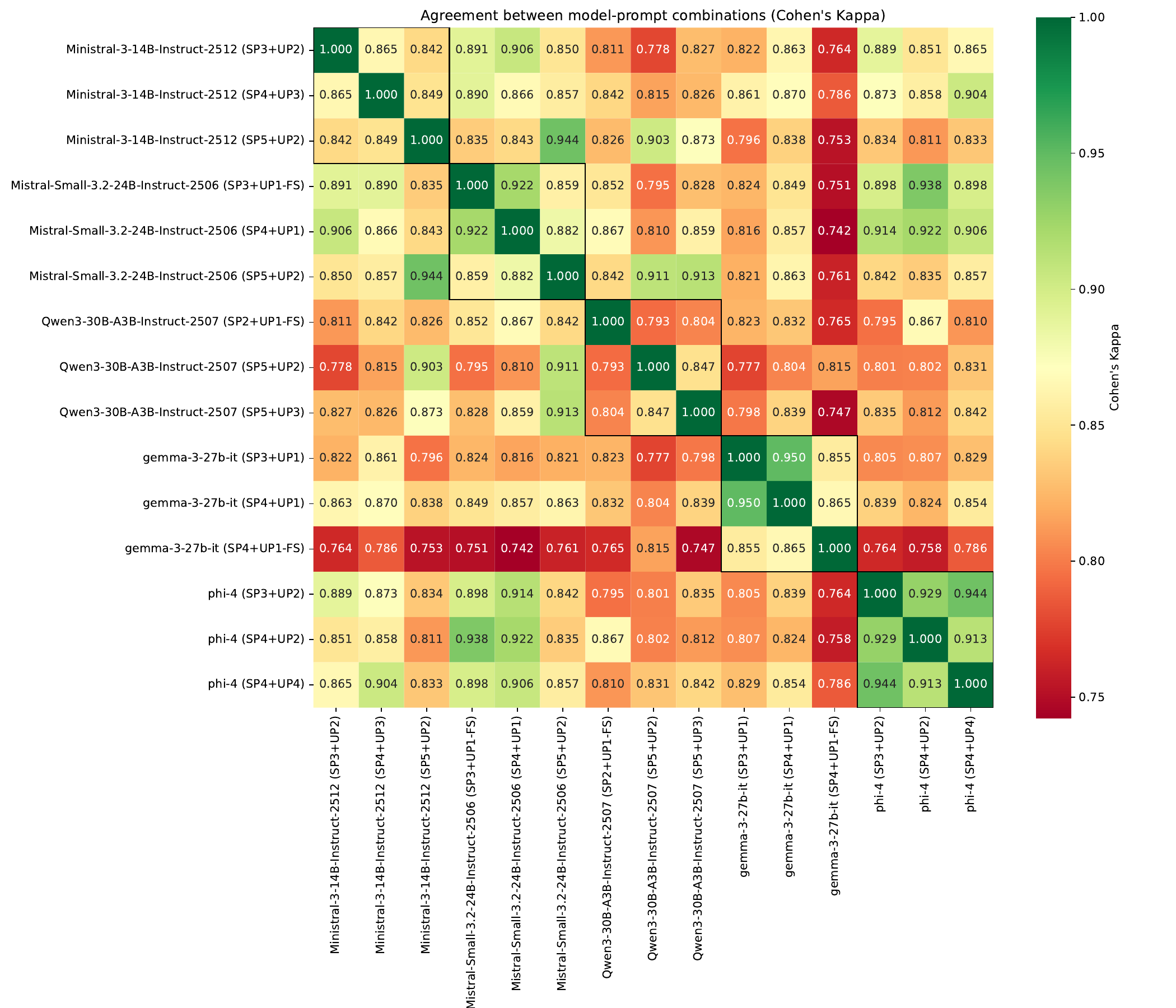}
    \caption{Cohen's Kappa between pairs of LLM annotators on the test set. Pairs with the same model are outlined in black.}
    \label{fig:app:model_agreement_heatmap}
\end{figure}

\subsubsection{Sensitivity to prompt and model choice}\label{app:sensitivity-test}
\begin{figure}[H]
    \centering
    \includegraphics[width=0.75\textwidth]{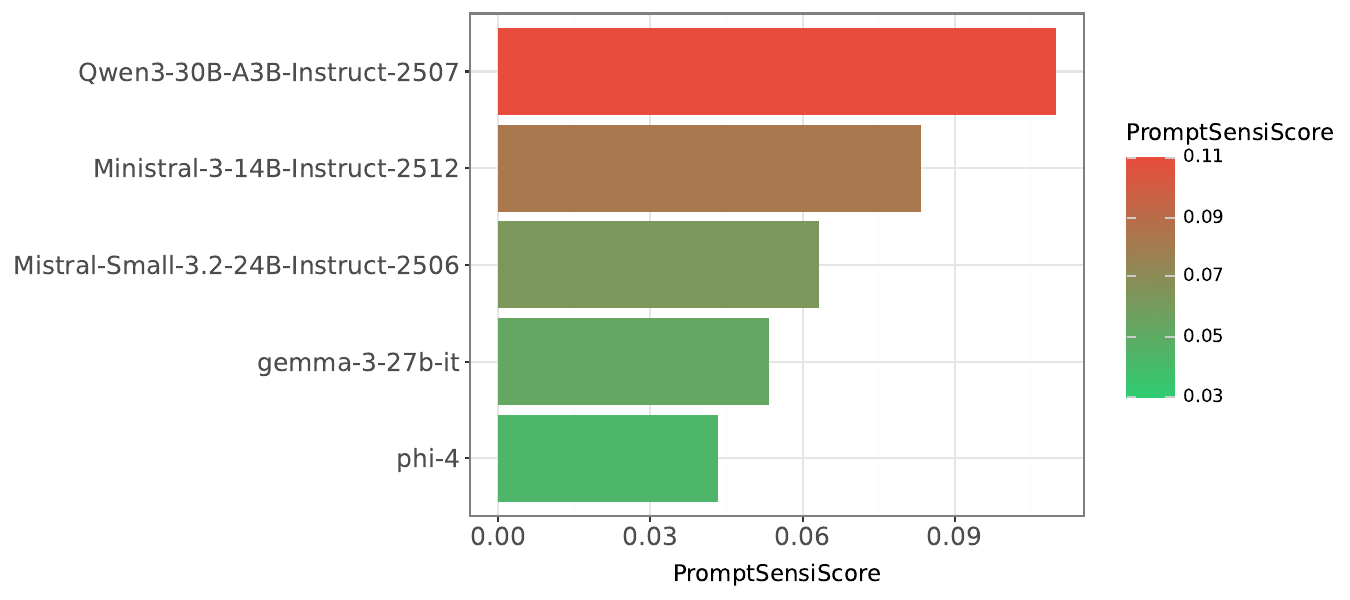}
    \caption{Instance-level prompt sensitivity across models. Prompt Sensitivity Scores (PSS) across 15 configurations (5 models $\times$ 3 prompt combinations each) on 200 test cases. Phi-4 shows lowest sensitivity (PSS=0.043), gemma-3-27b (PSS=0.053), Mistral-Small-24B (PSS=0.063), Ministral-14B (PSS=0.087), and Qwen3-30B highest (PSS=0.110). Green indicates robust, red indicates sensitive.}
    \label{fig:prompt_sensitivity}
\end{figure}

\begin{figure}[H]
    \centering
    \includegraphics[width=0.75\textwidth]{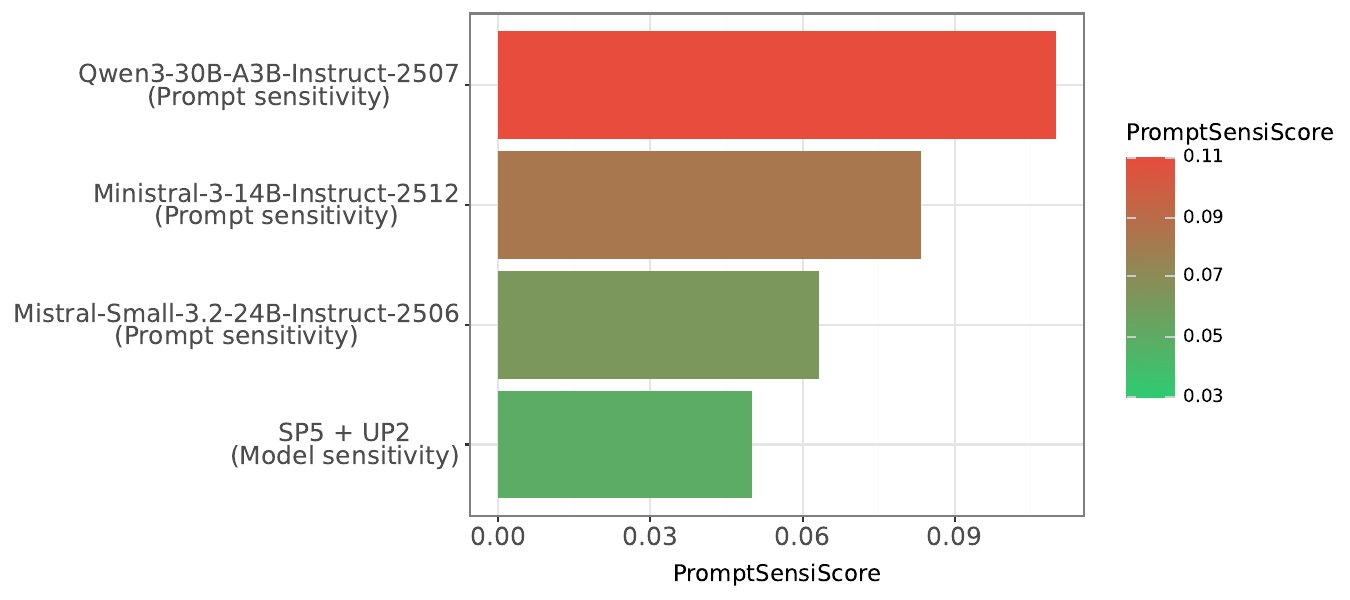}
    \caption{Model sensitivity versus prompt sensitivity for SP5+UP2. Model sensitivity (green, PSS=0.110) computed across three architectures (Ministral-14B, Mistral-Small-24B, Qwen3-30B) using identical SP5+UP2 prompts on 200 cases, versus prompt sensitivity within each model across three prompts (red/orange/green bars: PSS=0.110, 0.087, 0.063 respectively).}
    
    \label{fig:model_sensitivity}
\end{figure}

\subsubsection{Inter-class confusion}\label{app:test-interclass-confusion}

\begin{figure}[h]
    \centering
    \includegraphics[width=\linewidth]{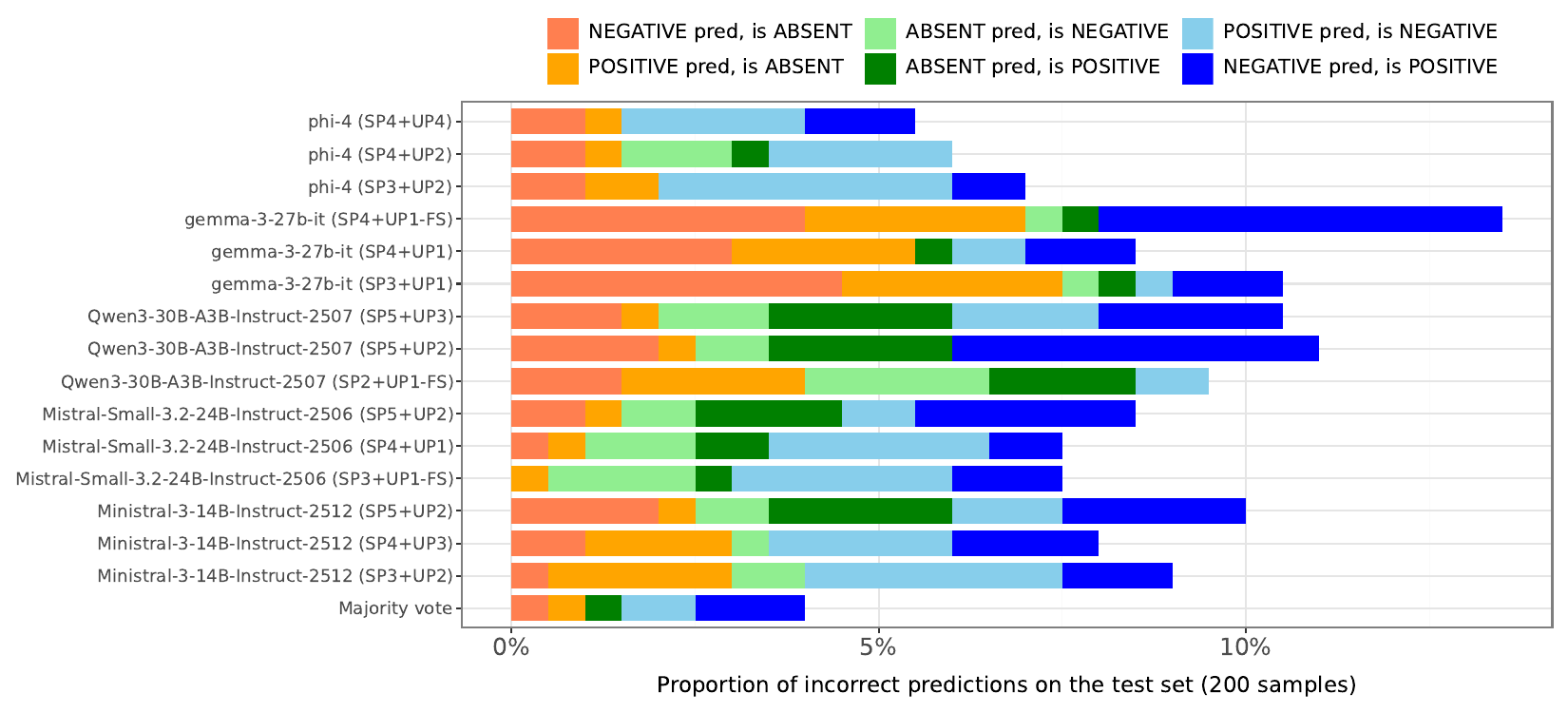}
    \caption{(Larger version of Figure~\ref{fig:credibility_pred_classes}) Mistakes made by individual LLMs and the ensemble (bottom), color-coded by class confusion.}
    \label{app:fig:credibility_pred_classes}
\end{figure}

\begin{figure}[h]
    \centering
    
\end{figure}

% Side-by-side figures
\begin{figure}
\centering
\begin{minipage}{.4\textwidth}
  \centering
  \includegraphics[height=4.2cm]{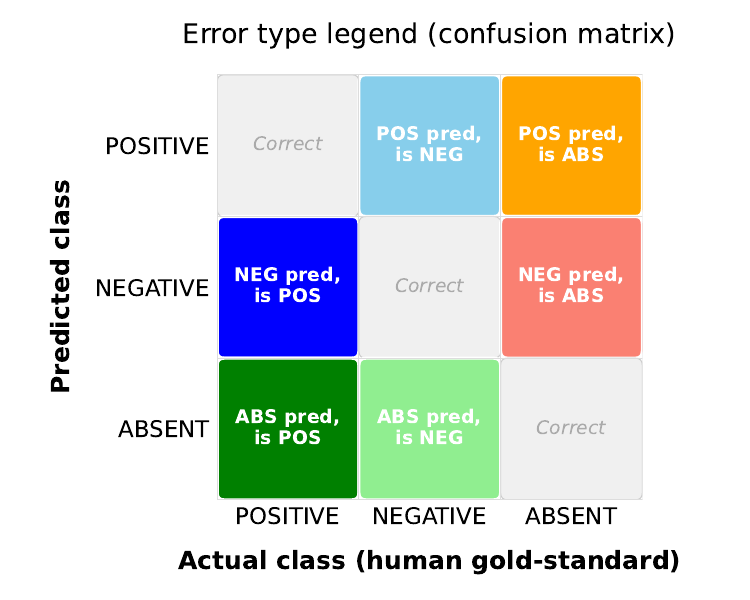}
    \captionof{figure}{Legend for Figure~\ref{app:fig:credibility_pred_classes}}
\end{minipage}%
\begin{minipage}{.6\textwidth}
  \centering
  \includegraphics[height=4.2cm]{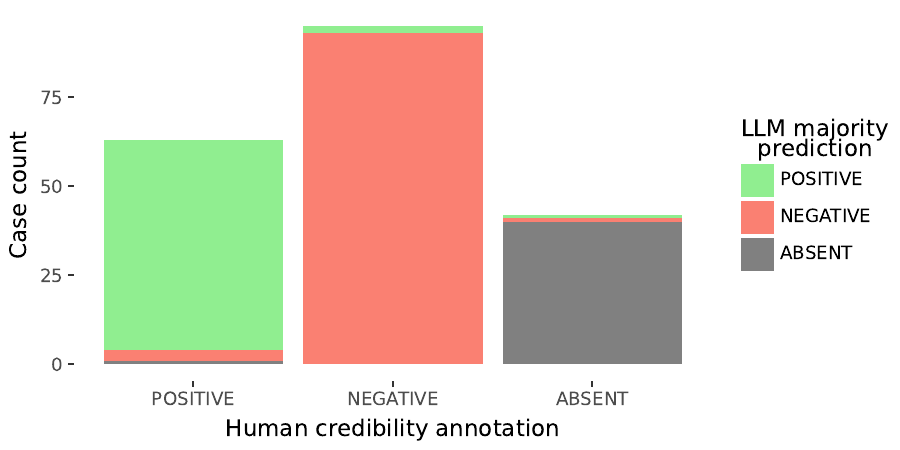}
  \captionof{figure}{LLM majority prediction (majority vote across ensemble of 15 LLMs) vs. gold standard human annotation.}
  \label{fig:test2}
\end{minipage}
\end{figure}

\subsubsection{Consistently misclassified cases}\label{app:misclassified-cases-test}

H1 was asked to rate mistake severity according to the following 3 categories:

\begin{itemize}[itemsep=0pt,topsep=0pt,parsep=0pt]
    \item[(A)] \textit{I consider this mistake to be an acceptable answer, as I hesitated myself between the LLM's prediction \& what I picked when annotating / this mistake is making me rethink/reconsider my own annotation}
    \item[(B)] \textit{mistake is understandable and not severe, but I would not expect a fellow domain expert who fully understands the codebook to make it}
    \item[(C)] \textit{mistake is severe / not acceptable / calls for an improvement of the model/prompt/etc.}
\end{itemize}

\noindent The two cases (ID 919 and 4317) which are most frequently misclassified across the LLM annotators  were rated as (A) by H1, and are shown in Table~\ref{tab:misclassified_examples_test}, along with a reflection by H1 and an LLM's reasoning.

\begin{casebox}[title={\textbf{Case ID 919 (test set)}}]
In January 2023, the Board granted a residence permit (Refugee Convention Status) to a male citizen of Afghanistan. Earlier in 2018, the Board had upheld the Danish Immigration Service's decision to refuse asylum, after which the Board reopened the case in 2022. The person in question had entered the country in 2015. The Refugee Appeals Board stated: "The applicant is an ethnic Hazara and a Shiite Muslim from [city A], [province], Afghanistan. The applicant was born and raised in [city B], Iran. The applicant has not been a member of any political or religious associations or organizations, nor has he been politically active in any other way. The applicant originally cited as his reason for seeking asylum that he feared being killed by the Taliban if he returned to Afghanistan. His family left Afghanistan in 1996 after a conflict with the Taliban, who had killed [several of the applicant's father's family members]. During the reopening of the case, the applicant cited as his reason for seeking asylum that, as a result of the change of power in Afghanistan and the current situation in the country, he would be at risk of persecution because he would be considered Westernized. The applicant further stated that he is of Hazara ethnicity and a Shiite Muslim.

With regard to the applicant's original reason for asylum, which is linked to a derived risk to the applicant due to his family's conflict with the Taliban, reference is made to the Refugee Board's decision of [autumn] 2018 and the reasons stated therein. The fact that the Taliban has now taken power in Afghanistan cannot lead to a different assessment. It follows from the above that this reason for asylum cannot form the basis for asylum under Section 7 of the Aliens Act. The conditions for Westernized persons and persons of Hazara ethnicity (and Shiite Muslims) and the risk assessment to be made in this connection in the event of a return to Afghanistan are described in the background material, inter alia, in sections 3.13 and 3.14 of EUAA, Country Guidance: Afghanistan, January 2023. The Refugee Appeals Board notes in this connection that the decisive factor in the asylum assessment in cases involving questions of so-called Westernization is whether, based on an overall assessment, the applicant has such experience of life in Afghanistan that he or she will be able to resume their life in the country in such a way that he or she will not attract the attention of the Taliban. In this regard, it should be noted that the applicant entered Denmark as an unaccompanied minor in [autumn] 2015, when he was [12-15 years old], and that he has therefore spent a significant part of his schooling in [middle school and high school] and his youth in Denmark, where he has lived as a Danish teenager, including the personal challenges he has explained in the lawyer's statement and during the board meeting. The applicant speaks fluent Danish, and the board meeting was conducted in Danish. Based on the applicant's appearance at the board meeting and the information provided, it must be assumed that the applicant has adopted a Western lifestyle. Since the applicant, according to his explanation, which the Board has taken as a basis, has never lived in Afghanistan, was born and raised in Iran, speaks Dari with an Iranian dialect, and has neither family nor other network in Afghanistan that could support and help him upon return, the Refugee Board finds – also taking into account that the applicant is of Hazara ethnicity (and a Shiite Muslim) – that, based on an overall assessment, it must be considered probable that the applicant will not be able to establish himself in Afghanistan in such a way that he does not attract the attention of the Taliban. Furthermore, and since the applicant is of Hazara ethnicity, as stated, the Refugee Appeals Board finds, after assessing the overall circumstances of the case, that the applicant has demonstrated that he would be at risk of persecution covered by section 7(1) of the Aliens Act if he returned to Afghanistan. The Refugee Board therefore grants the applicant a residence permit pursuant to section 7(1) of the Aliens Act. Reference number: Afgh/2023/8/MKTO.

\vspace{2mm}
\textit{Original case text:} \url{https://fln.dk/praksis/2023/februar/afgh20238/}

\hrulefill

\vspace{2mm}

\textbf{Annotator H1}: Q1: Credibility assessment present? \textbf{No} (Confidence: Low)

\textbf{Annotator H2}: Q1: Credibility assessment present? \textbf{No} (Confidence: Medium)

\vspace{2mm}
\textbf{15 LLM annotators}: Positive credibility assessment \textcolor{red}{\xmark}

\hrulefill

\vspace{2mm}

\textbf{\textit{H1's reflection:}} I think the issue could stem from the part around here: "\textit{Herefter, efter ansøgerens fremtræden under nævnsmødet og det i øvrigt oplyste må det lægges til grund, at ansøgeren har tilegnet sig en vestlig livsstil. Da ansøgeren samtidig efter sin forklaring, som nævnet har lagt til grund, aldrig har boet i Afghanistan, er født og opvokset i Iran, taler dari med iransk dialekt og hverken har familie eller øvrigt netværk i Afghanistan, der ville kunne støtte og hjælpe ham ved en tilbagevenden, finder Flygtningenævnet - også under hensyntagen til, at ansøgeren er hazara af etnicitet (og shiamuslim) - at det efter en samlet vurdering må anses for sandsynliggjort, at ansøgeren ikke vil kunne etablere sig i Afghanistan på en sådan måde, at han ikke påkalder sig Talibans opmærksomhed.}"

 First of all, there is the part about "lægges til grund" (usually used a lot for the credibility) that the claimant has acclaimed a Western lifestyle. I think for me the core of this case was the probability of the claimant to establish a life without attracting the attention of the Taliban and whether the person had acquired a Western enough lifestyle. The style of writing for me generally came down to both a future oriented judgment of what is Western enough or different enough to reach a threshold of potentially attracting attention and hence being at risk of persecution (more the direction of risk assessment), and less on whether the identity of the claimant can be accepted or past events described are accepted as the basis of the testimony, and I think therefore the core of the issue discussed in this case went for me into a different kind of assessment than classic, text-book credibility of "do we \textit{lægge til grund} what you are saying about yourself and your past experiences to use this as a basis to continue further assessment on". I can imagine that an LLM does not distinguish that necessarily as a categorically different assessment, and it might be very well that I went astray here on the human side. 
 
 Further, there is this second part that I copied in, about the accepting that the claimant never lived in Afghanistan, grew up in Iran etc. and reading this part I do think very much that this alone I would label as a credibility assessment, and I think that this is why I scored with low confidence on these cases, and absolutely can see how I as a human took one too many assumptions."

 \hrulefill

\vspace{2mm}

\textbf{\textit{Reasoning output of phi-4, prompted with SP4+UP4:}}

\vspace{\promptspacing}

The task is to determine whether a credibility assessment is made in the decision of the Danish Refugee Appeals Board (Flygtningenævnet) and, if so, to classify its sentiment as either positive, negative, or no credibility assessment. The decision involves a male national from Afghanistan who was granted a residence permit (K-status) in January 2023. The case involves assessing the risk of persecution upon return to Afghanistan, considering the applicant's background as an ethnic Hazara Shia Muslim, his upbringing in Iran, and his life in Denmark.

\vspace{\promptspacing}

\textbf{\#\#\# Step 2: Preliminary Judgment}

Upon reviewing the case text, it appears that the decision does involve a credibility assessment. The decision explicitly states that the applicant's explanation about not having lived in Afghanistan, being raised in Iran, and having a Western lifestyle is accepted ("nævnet har lagt til grund"). This acceptance of the applicant's account suggests a positive credibility assessment, as the core of his narrative is believed, allowing the risk assessment to proceed with his information.

\vspace{\promptspacing}

\textbf{\#\#\# Step 3: Critical Assessment}

To ensure accuracy, let's critically assess the decision:

\textbullet\ \textbf{Credibility Assessment Indicators}: The decision uses phrases like "nævnet har lagt til grund," indicating that the applicant's explanation is accepted as true. This aligns with the criteria for a positive credibility assessment.

\textbullet\ \textbf{Risk Assessment}: The decision proceeds to assess the risk of persecution based on the accepted account, considering the applicant's Western lifestyle and lack of ties in Afghanistan.

\textbullet\ \textbf{No Explicit Rejection}: There is no language indicating that the core of the applicant's account is rejected or deemed non-credible.

Given these points, the preliminary judgment of a positive credibility assessment seems consistent with the decision's content.

\vspace{\promptspacing}

\textbf{\#\#\# Step 4: Confirm Final Answer}

The final classification is \textbf{POSITIVE CREDIBILITY ASSESSMENT}. The decision explicitly accepts the core of the applicant's account, allowing the risk assessment to proceed with his information. The decision does not indicate any rejection of the core narrative, and the applicant's explanation is used as a basis for further analysis.

\vspace{\promptspacing}

\textbf{\#\#\# Step 5: Evaluate Confidence}

I am confident (90\%) in this analysis. The decision clearly uses language that indicates acceptance of the applicant's core narrative, which is a key indicator of a positive credibility assessment. The decision's structure and content align well with the criteria for this classification. The remaining uncertainty (10\%) stems from the complexity of legal language and the potential for nuanced interpretations, but the evidence strongly supports the conclusion.

\end{casebox}

\begin{casebox}[title={\textbf{Case ID 4317 (test set)}}]
    In November 2018, the Board upheld the Danish Immigration Service’s decision regarding a male national from an unknown country of origin. He entered the country in 2015.
The Refugee Board stated:
“The applicant is an ethnic Arab and a Shiite Muslim from Al Jahra, Kuwait. The applicant has not been a member of any political or religious associations or organizations, nor has he been politically active in any other capacity. As grounds for asylum, the applicant has stated that he fears being imprisoned by the Kuwaiti authorities upon return to his home country due to his illegal departure. The applicant also fears the general conditions for Bidoons. In support of his asylum claim, the applicant has stated that he is an unregistered Bidoon and therefore has no right to a birth certificate, an ID card, schooling, or health care in Kuwait. The applicant has further stated that his uncle, [A], participated in a demonstration [in early] 2014. [In early] 2014, [A] was arrested at the family’s shared residence and was subsequently imprisoned until [spring] 2014. After Bader’s departure, the Kuwaiti authorities visited the applicant’s residence an unknown number of times, causing the applicant’s family to feel harassed. Against this background, the applicant left Kuwait in mid-[fall] 2015 together with his family. In the mother’s case, the Refugee Board has reached the following decision: “The Refugee Board finds that it cannot rely on the applicant’s statement regarding her identity, her nationality, or her grounds for asylum. In this connection, it is noted that the applicant has demonstrated limited knowledge of her home area, where she claims to have lived her entire life. For instance, she has been unable to identify street addresses in her neighborhood other than her own, nor is she familiar with the location of her father’s grave in relation to her residence. The documents submitted by the applicant do not lead to a different assessment of the credibility of her statement. Furthermore, she has provided conflicting accounts regarding visits to the authorities following her brother’s release, whose asylum claim was, incidentally, rejected by the Refugee Board in the fall of 2018. Against this background, the Board does not find that the applicant has established a likelihood that, upon return to her home country, she would be at a concrete or individual risk of persecution or abuse covered by Section 7 of the Aliens Act. It is added that the applicant has not reported having had conflicts in other countries, including not in Iraq. It is noted that the general conditions for Bidoon in Kuwait, viewed in isolation, do not constitute grounds for asylum, and that the sanctions for illegal departure, based on the available background information, are not considered disproportionate. The Refugee Board therefore upholds the Danish Immigration Service’s decision.” Since the applicant’s grounds for asylum correspond to those stated by his mother, and since the circumstances included in the decision regarding the applicant’s identity, including nationality, are identical to those of his mother, the Refugee Board, with reference to this decision, also finds that this applicant has not established a likelihood that, upon return to his home country, he would be at a concrete and individual risk of persecution or abuse covered by Section 7 of the Aliens Act. The Refugee Board therefore upholds the Danish Immigration Service’s decision.” Hjem/2018/88/CHA

\vspace{2mm}
    \textit{Original case text:} \url{https://fln.dk/praksis/2019/april/hjem201888/}

    \hrulefill
    
    \textbf{Annotator H1}: Q1: Credibility assessment present? \textbf{No} (Confidence: Low)
    
    \textbf{Annotator H2}: Q1: Credibility assessment present? \textbf{No} (Confidence: Low) 

    \vspace{2mm}
    
    \textbf{14 LLM annotators}: Negative credibility assessment \textcolor{red}{\xmark}

    \textbf{1 LLM annotator}: No credibility assessment \textcolor{green}{\cmark}

    \hrulefill

\vspace{2mm}

    \textbf{H1's reflection for case 4317}:
This might also be confusing to other human annotators and I am actually inclined to think that I should have labeled this differently (hence low confidence): The reasoning expands upon the credibility of the mother's explanation, and rejects the mother's testimony due to credibility issues. We had originally thought to exclude references to credibility assessments of family members, in cases where the claimant in the case to be annotated was only rejected or accepted based on the credibility of the family member (as this entails no credibility assessment explicitly on the claimant). However, in this particular case, I would say that they almost reject the claimant because they have the same motive as the mother, who is seen as not credible in her motive, i.e. I actually think one could argue that the claimant themselves is also perceived as non-credible. And from an LLM perspective I think there is so much language about credibility, that I am super interested in which combination actually ruled against this being credibility.

    \hrulefill

\vspace{2mm}

\textbf{\textit{Reasoning output of phi-4, prompted with SP4+UP4:}}

\vspace{\promptspacing}

\textbf{\#\#\# Step 1: Clarify Understanding of the Task and the Case}

The task is to analyze a decision from the Danish Refugee Appeals Board (Flygtningenævnet) to determine if a credibility assessment was made and, if so, its sentiment. The decision involves a male applicant from Kuwait, who claims asylum based on fears of persecution due to his status as an "unregistered" Bidoon and his family's experiences with Kuwaiti authorities. The decision must be classified into one of three categories: NO CREDIBILITY ASSESSMENT, POSITIVE CREDIBILITY ASSESSMENT, or NEGATIVE CREDIBILITY ASSESSMENT.

\vspace{\promptspacing}

\textbf{\#\#\# Step 2: Preliminary Judgment}

Upon reviewing the case text, it is evident that the Flygtningenævnet explicitly states that they "find not to be able to base the applicant's explanation regarding his identity, nationality, or asylum motive." This indicates a direct assessment of the applicant's credibility. The decision mentions inconsistencies and a lack of detailed knowledge about his home area, which further supports the conclusion that the core of the applicant's account is not accepted. Therefore, the preliminary judgment is that there is a NEGATIVE CREDIBILITY ASSESSMENT.

\vspace{\promptspacing}

\textbf{\#\#\# Step 3: Critical Assessment}

To ensure accuracy, let's critically assess the decision:

\textbullet\ \textbf{Explicit Statements}: The decision explicitly states that the applicant's explanations about identity, nationality, and asylum motive cannot be accepted. This is a clear indication of a credibility assessment.

\textbullet\ \textbf{Inconsistencies and Lack of Knowledge}: The decision highlights the applicant's limited knowledge of his home area and inconsistencies in his account, which are reasons for rejecting the core of his narrative.

\textbullet\ \textbf{Peripheral Facts}: While some peripheral facts might be accepted, the core elements necessary for asylum are not believed, aligning with a NEGATIVE CREDIBILITY ASSESSMENT.

\vspace{\promptspacing}

\textbf{\#\#\# Step 4: Confirm Final Answer}

The final answer is NEGATIVE CREDIBILITY ASSESSMENT. The decision explicitly rejects the core of the applicant's account, citing inconsistencies and a lack of detailed knowledge, which are critical for establishing credibility.

\vspace{\promptspacing}

\textbf{\#\#\# Step 5: Evaluate Confidence}

I am 95\% confident in this analysis. The decision clearly articulates reasons for rejecting the applicant's core narrative, which aligns with the criteria for a \textbf{NEGATIVE CREDIBILITY ASSESSMENT}. The explicit language used in the decision leaves little room for alternative interpretations regarding the credibility assessment. The remaining 5\% uncertainty accounts for any potential nuances in the interpretation of the decision's language that might not be immediately apparent.

\end{casebox}
  \captionof{table}{
    Translation of two case texts from the test set (ID 919 and 4317) which were misclassified by all or almost all 15 LLM annotators, and the corresponding annotation independently assigned by the two domain experts. Under each case text, we include the reasoning output of the best-performing LLM annotator on the test set (bold added for clarity). Case text translated with DeepL.com (free version).
  }\label{tab:misclassified_examples_test}

\end{document}